\documentclass[11pt]{article}
\usepackage[final]{acl}
\usepackage{times}
\usepackage{latexsym}
\usepackage[T1]{fontenc}
\usepackage[utf8]{inputenc}
\usepackage{microtype}
\usepackage{inconsolata}
\usepackage{graphicx}
\usepackage{amsmath}
\usepackage{subcaption}
\usepackage{float}
\usepackage{booktabs}
\usepackage{multirow}
\usepackage{lineno}
\usepackage{tcolorbox}
\tcbuselibrary{breakable}
\usepackage{caption}
\usepackage{tcolorbox}
\usepackage{pifont}
\usepackage{xcolor}
\usepackage{makecell}
\usepackage{booktabs}
\usepackage{multirow}
\usepackage{graphicx}
\usepackage{xcolor}
\usepackage{tcolorbox}
\usepackage{colortbl}
\usepackage{tikz}
\usepackage{pgfplots}
\usepackage{subcaption} 
\pgfplotsset{compat=1.18}
\definecolor{colorBase}{RGB}{149, 205, 216} 
\definecolor{colorSFT}{RGB}{114, 183, 219} 
\definecolor{colorRL}{RGB}{102, 149, 214}
\usepackage{caption}
\usepackage{algorithm}
\usepackage{algpseudocode}
\usepackage{booktabs}
\usepackage{authblk}
\usepackage{colortbl} 
\usepackage[table]{xcolor} 
\definecolor{highlightrow}{gray}{0.94} 
\definecolor{softblue}{rgb}{0.90, 0.95, 1.0}
\usepackage{makecell} 

\usepackage{pifont} 
\definecolor{darkgreen}{RGB}{0,160,0}
\newcommand{\notcheckmark}{\textcolor{black}{\bcmark\kern-1.1ex\raisebox{.7ex}{\rotatebox[origin=c]{125}{--}}}\color{black}}
\newcommand{\bcmark}{\color{blue}{\ding{51}}}%
\newcommand{\cmark}{\color{darkgreen}{\ding{51}}}%
\newcommand{\xmark}{\color{red}{\ding{55}}}%
\title{ChartVerse: Scaling Chart Reasoning via Reliable Programmatic Synthesis from Scratch}

\makeatletter
\renewcommand\AB@affilsepx{, \protect\Affilfont}
\makeatother

\author[1,2,*]{\textbf{Zheng~Liu}}
\author[1,3,*]{\textbf{Honglin Lin}}
\author[2,4,5,†]{
\textbf{Chonghan Qin}$^{1}$, 
\textbf{Xiaoyang Wang}$^{1}$, 
\textbf{Xin Gao}$^{1}$, 
\textbf{Yu Li}$^{1}$, \\
\textbf{Mengzhang Cai}$^{1}$,
\textbf{Yun Zhu}$^{1}$,
\textbf{Zhanping Zhong}$^{1}$, 
\textbf{Qizhi Pei}$^{1}$, 
\textbf{Zhuoshi Pan}$^{1}$,
\textbf{Xiaoran Shang}$^{1}$, \\
\textbf{Conghui He}$^{1}$, 
\textbf{Bin Cui}$^{2,5}$,
\textbf{Wentao Zhang}
}
\author[1,†]{\textbf{Lijun~Wu}}

\affil[1]{Shanghai AI Laboratory} \affil[2]{Peking University, \protect\\  \textsuperscript{3}Shanghai Jiao Tong University, \textsuperscript{4}Zhongguancun Academy, \protect\\ \textsuperscript{5}Beijing Key Laboratory of Software and Hardware Cooperative Artificial Intelligence Systems} 

\begin{document}
\maketitle

\let\thefootnote\relax
\footnotetext{* Equal contribution.}
\footnotetext{† Corresponding authors.}

\begin{abstract}

Chart reasoning is a critical capability for Vision Language Models (VLMs). However, the development of open-source models is severely hindered by the lack of high-quality training data. Existing datasets suffer from a dual challenge: synthetic charts are often simplistic and repetitive, while the associated QA pairs are prone to hallucinations and lack the reasoning depth required for complex tasks.
To bridge this gap, we propose \textbf{ChartVerse}, a scalable framework designed to synthesize complex charts and reliable reasoning data from scratch. (1) To address the bottleneck of simple patterns, we first introduce \textit{Rollout Posterior Entropy (RPE)}, a novel metric that quantifies chart complexity. Guided by RPE, we develop \textit{complexity-aware chart coder} to autonomously synthesize diverse, high-complexity charts via executable programs. (2) To guarantee reasoning rigor, we develop \textit{truth-anchored inverse QA synthesis}. Diverging from standard generation, we adopt an answer-first paradigm: we extract deterministic answers directly from the source code, generate questions conditional on these anchors, and enforce strict consistency verification. To further elevate difficulty and reasoning depth, we filter samples based on model fail-rate and distill high-quality Chain-of-Thought (CoT) reasoning.
We curate ChartVerse-SFT-600K and ChartVerse-RL-40K using Qwen3-VL-30B-A3B-Thinking as the teacher. Experimental results demonstrate that ChartVerse-8B achieves state-of-the-art performance, notably surpassing its teacher and rivaling the stronger Qwen3-VL-32B-Thinking. We release our code, model weights, and datasets in \href{https://chartverse.github.io}{https://chartverse.github.io}.

\end{abstract}
\section{Introduction}
\label{sec:intro}

Recent Vision Language Models (VLMs)~\citep{bai2025qwen3vltechnicalreport,openmmreasoner,gpt5,gemini3} exhibit strong reasoning capabilities, yet their performance on chart reasoning remains far from satisfactory. A central limitation lies in the lack of high-quality data that simultaneously achieves  scale, diversity, and complexity.
For example, manually curated datasets~\citep{masry2022chartqabenchmarkquestionanswering,methani2020plotqareasoningscientificplots,xiao2025visual,INTENT,REFINE} provide realistic chart distributions but are severely limited by annotation cost. Synthetic approaches—ranging from template-based rendering~\citep{kafle2018dvqaunderstandingdatavisualizations,han2023chartllamamultimodalllmchart,xiao2025prompt,STABLE,MELT,HINT,REFINE} to recent code-driven pipelines~\citep{he2025distillvisualchartreasoning,zhao2025chartcoderadvancingmultimodallarge,kondic2025chartgenscalingchartunderstanding,INTENT,OFFSET,HABIT,ReTrack}—offer scalability but typically rely on fixed templates or seed charts, resulting in limited visual diversity and poor coverage of real-world, long-tail chart patterns. Although proprietary systems~\citep{yang2025scalingtextrichimageunderstanding,ReTrack,HABIT,OFFSET} can produce visually rich charts, their dependence on closed commercial APIs prevents large-scale, open-source adoption.

Beyond visual diversity, the quality of reasoning supervision presents a second major challenge. Existing chart Question-Answering (QA) datasets are constrained in both difficulty and reliability. Visually simplistic charts inherently limit question complexity, making it difficult to construct multi-step reasoning problems. Moreover, answer correctness is also hard to guarantee at scale: manual verification~\citep{masry2022chartqabenchmarkquestionanswering} is infeasible, while LLM-generated QA pairs~\citep{xu2025chartm3multistagecodedrivenpipeline,yang2025scalingtextrichimageunderstanding,chen2025chartr1chainofthoughtsupervisionreinforcement,li2026whatsmissingscreentoactionuiintheloop,ye2025tableqa,guo2026rethinking,wu2025table,yang2026infactdiagnosticbenchmarkinduced,ZHANG2026112674} frequently suffer from hallucinations and numerical errors. As a result, current pipelines struggle to provide large-scale, high-fidelity chart reasoning data with rigorous and error-free supervision for model training.

\begin{table*}[htbp]
\setlength{\tabcolsep}{5pt}
  \centering
  \caption{Comparison of ChartVerse-SFT-600K with existing chart datasets. }
    \resizebox{1.0\textwidth}{!}{
    \begin{tabular}{lcccccccccc}
    \toprule
    \multirow{3}{*}{\textbf{Datasets}} & \multicolumn{3}{c}{\textbf{Chart Properties}} & \multicolumn{4}{c}{\textbf{Q\&A Properties}} & \multicolumn{3}{c}{\textbf{Complexity \& Diversity}} \\
    \cmidrule(lr){2-4} \cmidrule(lr){5-8}  \cmidrule(lr){9-11} 
    & \makecell{ Chart\\Count} & \makecell{Data Source} & \makecell{Textual\\Data} & \makecell{QA\\Count} & 
    \makecell{Total\\Tokens} &\makecell{Answer\\Accuracy} & \makecell{Reasoning\\Data} & \makecell{Rollout Posterior\\Entropy} & \makecell{Color\\Entropy} & \makecell{Semantic Emb\\Spread}\\
    \midrule
    ChartQA~\citep{masry2022chartqabenchmarkquestionanswering} & 18K & Real-world, Synthetic & Table & 28K & 100K & \xmark & \xmark & 0.26 & 2.03 & 0.37 \\
    PlotQA~\citep{methani2020plotqareasoningscientificplots} & 156K & Real-world, Synthetic & Table & 20M & 117M & \xmark & \xmark & 0.21 & 0.83 & 0.30\\
    FigureQA~\citep{kahou2018figureqaannotatedfiguredataset} & 100K & Synthetic & - & 1.3M & 2.6M  & \xmark & \xmark & 0.25 & 1.04 & 0.24\\
    ReachQA~\citep{he2025distillvisualchartreasoning} & 3K & Synthetic & Code & 20K & 1.2M  & \xmark & \cmark & 0.29 & 1.91 & 0.47\\
    ECD~\citep{yang2025effectivetrainingdatasynthesis} & 10K & Synthetic & Code & 321K & 18.6M & \xmark & \cmark & 0.31 & 2.13 & 0.40 \\
    CoSyn~\citep{yang2025scalingtextrichimageunderstanding} & 116K & Synthetic & Code &  1.1M &  5.4M & \xmark & \xmark & 0.35 & 1.73 & 0.54  \\
    START~\citep{liu2025startspatialtextuallearning} & 400K & Synthetic & Code &  400K & 143M & \xmark & \xmark & 0.33 & 1.63 & 0.27\\
    ChartGen~\citep{kondic2025chartgenscalingchartunderstanding} & 216K & Synthetic & Code &  - & - & \xmark & \xmark & 0.30 & 1.31 & 0.39 \\
    ChartCoder~\citep{zhao2025chartcoderadvancingmultimodallarge} & 163K & Synthetic & Code &  - & - & \xmark & \xmark & 0.33 & 1.48 & 0.38 \\
    \rowcolor{softblue}
    ChartVerse-SFT-600K(Ours) & 412K & Synthetic & Code &  603K & 3.9B & \cmark & \cmark & 0.44 & 3.17 & 0.51 \\
    \bottomrule
    \end{tabular}%
    }
\label{tab:chart_cmp}
\end{table*}%

To address the above challenges, we propose \textbf{ChartVerse}, a scalable, code-driven framework for chart reasoning data synthesis. ChartVerse is designed to jointly improve visual complexity, distributional diversity, and reasoning reliability, enabling the construction of large-scale, challenging chart QA data without relying on fixed templates or manual verification.

We first introduce \textbf{Rollout Posterior Entropy (RPE)}, a metric for estimating intrinsic chart difficulty. Given a chart image, we repeatedly prompt an existing VLM to parse the image into executable chart code and measure the consistency across multiple rollouts. Charts that admit diverse, inconsistent code reconstructions exhibit higher posterior entropy, indicating greater visual and semantic ambiguity. RPE thus provides an objective, model-agnostic signal for identifying challenging charts and guiding data selection.

Using RPE, we first filter existing chart corpora to construct a high-difficulty seed dataset, which is then used to train a \textbf{complexity-aware chart coder}. Unlike prior template-based or seed-conditioned approaches, our chart coder generates chart code from scratch via high-temperature sampling, enabling broad exploration of the long-tail chart distribution. Through an iterative loop of generation, difficulty filtering, and retraining, the chart coder progressively improves both code quality and visual complexity, producing diverse, realistic charts.

Further, to ensure both correctness and reasoning difficulty for QA, ChartVerse adopts a \textbf{truth-anchored inverse QA synthesis} pipeline. Ground-truth answers are first computed directly from chart code, guaranteeing numerical and semantic accuracy. Questions are then reverse-synthesized conditioned on these answers. To control difficulty, we evaluate the resulting QA pairs using a strong VLM and retain samples with non-trivial fail-rate, ensuring that the final dataset emphasizes genuinely challenging reasoning cases.

Leveraging Qwen3-VL-30B-A3B-Thinking as the teacher model, we finally distill high-quality chain-of-thought (CoT) supervision and curate two datasets: ChartVerse-SFT-600K and ChartVerse-RL-40K. Initializing from the Qwen3-VL Instruct series~\citep{bai2025qwen3vltechnicalreport}, we train ChartVerse models at 2B, 4B, and 8B scales. Despite their smaller sizes, these models achieve state-of-the-art (SOTA) performance on chart reasoning benchmarks. Notably, ChartVerse-4B surpasses Qwen3-VL-8B-Thinking, while ChartVerse-8B even outperforms its 30B teacher, approaching the performance of Qwen3-VL-32B-Thinking.

\begin{figure*}[!t]
    \centering
    \includegraphics[width=15cm]{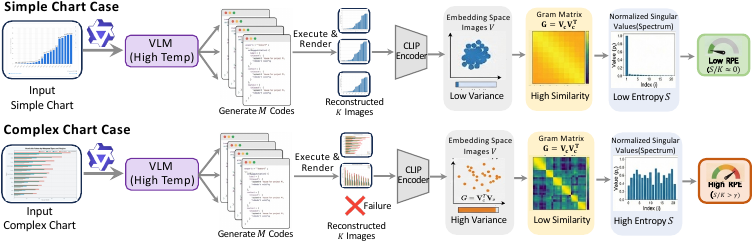}
    \caption{Illustration of Rollout Posterior Entropy. We quantify complexity via generative stability: simple charts yield consistent reconstructions (low RPE), while complex charts result in divergent outcomes (high RPE).}
    \label{fig:rpe}
\end{figure*}
\section{Related Work}
\paragraph{Chart VLMs.}
Leading proprietary models~\citep{gpt5,gemini3,Claude3.5,bai2025qwen3vltechnicalreport} have demonstrated exceptional reasoning capabilities in complex chart understanding. In the open-source domain, early methods like ChartLlama~\citep{han2023chartllamamultimodalllmchart}, ChartGemma~\citep{masry2024chartgemmavisualinstructiontuningchart}, and EvoChart~\citep{huang2025evochartbenchmarkselftrainingapproach} primarily focused on visual instruction tuning to handle fundamental structural extraction tasks like OCR and element counting. Recent works have transitioned towards reasoning-centric designs. TinyChart~\citep{zhang2024tinychartefficientchartunderstanding} employs template-based Program-of-Thought synthesis. 
ChartReasoner~\citep{jia2025chartreasonercodedrivenmodalitybridging} and Chart-R1~\citep{chen2025chartr1chainofthoughtsupervisionreinforcement} integrate reinforcement learning with supervised finetuning, directly incentivizing the model to generate robust CoT paths for complex problem-solving.

\paragraph{Chart Image Synthesis.}
Conventional approaches~\citep{kafle2018dvqaunderstandingdatavisualizations,han2023chartllamamultimodalllmchart,HINT,MELT,STABLE} rely on rule-based rendering engines to visualize data tables. These methods often yield charts with rigid layouts and limited diversity. Recent works~\cite{zhao2025chartcoderadvancingmultimodallarge, he2025distillvisualchartreasoning, xu2025chartm3multistagecodedrivenpipeline, jiang2025chartcocaselfimprovingchartunderstanding, kondic2025chartgenscalingchartunderstanding, yang2025effectivetrainingdatasynthesis,yuan2025understanding,ding2025arm,ding2025mmifenginemultimodalinstructionfollowing,zhao2025omnialign,huang2026vision,zeng2026vision,han2026unicorn} leverage LLMs to generate executable code or augment fixed visual seeds. Despite offering better flexibility, these methods remain constrained by finite and simple templates. The charts exhibit trivial topologies that are insufficient to elicit the complex reasoning capabilities of VLMs. Methods like Cosyn~\citep{yang2025scalingtextrichimageunderstanding,liuflare,lin2026mmfinereason} employ proprietary model Claude-3.5-Sonnet for end-to-end synthesis to avoid template restrictions. However, the dependence on commercial APIs introduces prohibitive costs and scalability bottlenecks.
\paragraph{Chart QA Construction.}
High-quality QA pairs is pivotal for chart understanding. While human annotation~\citep{masry2022chartqabenchmarkquestionanswering,methani2020plotqareasoningscientificplots,liu2025uniform} guarantees accuracy, its unscalability has necessitated a shift towards automated generation. Recent approaches~\citep{yang2025scalingtextrichimageunderstanding, chen2025chartr1chainofthoughtsupervisionreinforcement, masry2024chartinstructinstructiontuningchart,lin2025scalingcodeassistedchainofthoughtsinstructions,liu2024mmcadvancingmultimodalchart,meng2024chartassisstantuniversalchartmultimodal,liu2025synthvlm,liu2025fusion,niu2025mineru2,li2025taco,li2025miv,li2025cama,li2025catp} typically prompt leading VLMs to synthesize QA pairs from images, with some works~\citep{he2025distillvisualchartreasoning,lin2026scientific,li2025faithact,zhang2025critic,li2026planviz,li2025chemvlm} leveraging auxiliary code. Despite their popularity, these methods are constrained by the stochastic nature of LLMs, which leads to unavoidable hallucinations and inaccuracies. Lacking ground truth for verification, these methods predominantly yield uncalibrated, simplistic questions and lack multi-step CoT, accurate answers.

\section{Methodology}
\begin{figure*}[!t]
    \centering
    \includegraphics[width=1.0\linewidth]{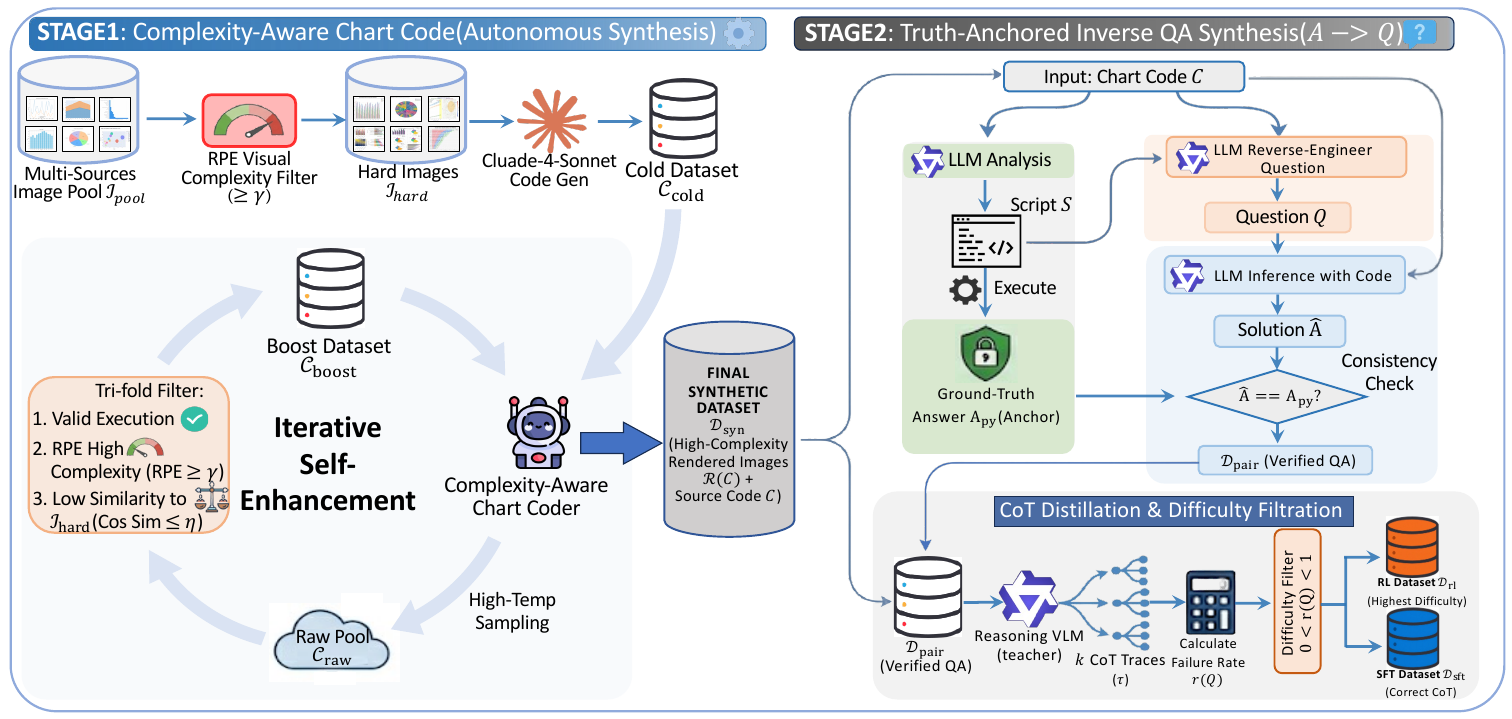}
    \caption{Overview of ChartVerse. Stage 1 trains a complexity-aware chart coder via RPE-guided iterative self-enhancement to synthesize diverse, high-complexity chart codes. Stage 2 generates verifiable QA pairs using truth-anchored inverse synthesis and CoT distillation, followed by failure-rate filtering to guarantee difficulty.}
    \label{fig:pipeline}
\end{figure*}

In this section, we introduce our ChartVerse framework for scalable chart reasoning data synthesis. We first introduce a difficulty metric to quantify intrinsic chart complexity, which guides data selection and generation. Based on this metric, we develop a complexity-aware chart coder and a truth-anchored inverse QA synthesis pipeline to produce challenging and reliable training data. Figure~\ref{fig:pipeline} shows the complete process.

\subsection{Quantifying Intrinsic Chart Complexity}
\label{sec:RPE}
A prerequisite for training a robust chart generator is the identification and filtration of high-complexity samples. However, distinguishing complexity via direct pixel analysis is intractable, as visual density does not correlate with structural difficulty. When humans interpret a complex chart, they often form diverse understandings and struggle to capture every detail. This implies that the decoding space for complex visual data is variable, making consistent reproduction difficult.

Guided by this insight, we propose Rollout Posterior Entropy (RPE) to measure the variability. Specifically, we employ a VLM to generate executable code for charts as a proxy for structural interpretation. As illustrated in Figure~\ref{fig:rpe}, 
simple charts yield consistent code that renders into identical images, whereas complex charts induce high instability, resulting in divergent reconstructions. 


\paragraph{Modeling Structural Variability.}
Given a chart, we utilize Qwen3-VL-2B-Thinking to generate plotting code 8 times with the temperature set to 1.0. We then execute these codes to render reconstructed images. Let $K$  denote the number of successfully executed samples. We utilize CLIP~\citep{radford2021learningtransferablevisualmodels} to extract $1 \times d$ dimensional features for each image and stack them into an image matrix $V \in \mathbf{R}^{K \times d}$. 

To quantify the variability, we first center $V$ to remove translational shifts and compute the Gram matrix $G$ to model pairwise similarities:
{
\begin{equation}
    V_{c} = \left(I - \frac{1}{K}\mathbf{1}\mathbf{1}^{T}\right) V , \quad G = V_{c}V_{c}^{T}.
\end{equation}
}

Next, we calculate the singular values $\sigma_i$ of $G$. We normalize these values and compute the spectral entropy to measure the degree of variation. 
{
\begin{equation}
    p_i = \sigma_i / \sum \sigma_j, \quad S = -\sum_{i=1}^{K} p_{i} \log p_{i}.
\end{equation}
}

Since execution failures inherently signal difficulty, we normalize $S$ by the valid count $K$ to obtain final RPE:
{
\begin{equation}
    RPE = \frac{S}{K}.
\end{equation}
}

Higher $RPE$ denotes greater reproduction difficulty, reflecting intrinsic chart complexity.

\paragraph{Diagnosing Existing Datasets.}
\label{ssec:Diagnosing}
Leveraging RPE, we evaluate the complexity of mainstream datasets. As shown in Table~\ref{tab:chart_cmp}, nearly all existing datasets exhibit low RPE scores, indicating that such charts are trivial. To drive further progress, it is critical to synthesize data that transcends simple visual variation, offering the complexity required to challenge the consistency of modern VLMs.

\subsection{Complexity-Aware Chart Coder}

RPE suggests that executable code is a high-fidelity proxy of chart structure, we therefore synthesize complex charts via code and propose a complexity-aware chart coder that can autonomously generate non-standard, high-complexity charts from scratch.

\paragraph{Difficulty-Filtered Cold Start.}
We first construct a high-quality cold-start seed code $\mathcal{C}_{cold}$. We aggregate chart images from existing datasets in Table~\ref{tab:chart_cmp} to form an image pool $\mathcal{I}_{pool}$ and apply RPE to quantify visual complexity. Based on the statistics in Table~\ref{tab:chart_cmp}, we set an RPE threshold of 0.4 to retain only high RPE samples:
{
\begin{equation}
\label{eq:ihard}
\mathcal{I}_{hard}=\{x\in\mathcal{I}_{pool}\mid RPE(x)\ge 0.4\},
\end{equation}
}

Since real-world charts in $\mathcal{I}_{hard}$ lack ground-truth source code, we employ Claude-4-Sonnet to infer candidate code $C(x)$ for each image $x\in\mathcal{I}_{hard}$. We discard samples that trigger execution errors, obtaining the cold-start training set:
{
\begin{equation}
\label{eq:ccold}
\mathcal{C}_{cold}=\{C(x)\mid x\in\mathcal{I}_{hard}\}.
\end{equation}
}

Ultimately, we curate a cold-start set $\mathcal{C}_{cold}$ comprising 60k high-quality samples, which serves as the foundational corpus for our chart coder.

Unlike existing methods that rely on code as rigid templates, we train a strong coder LLM, Qwen2.5-Coder-7B~\citep{hui2024qwen25codertechnicalreport}, to function as a flexible generative model.
We structure each training instance by pairing a concise system instruction $\mathcal{T}$ (see Prompt~\ref{tab:system_coder_prompt}) as the model input with the chart code sequence $C=(c_1,c_2,\dots,c_L) \in \mathcal{C}_{cold}$ as the target answer. The model is optimized using the standard cross-entropy loss to internalize the structural and logic patterns of executable chart codes. During inference, by applying a high sampling temperature, the coder leverages its aligned internal knowledge and learned patterns to generate diverse chart codes. This approach enables the model to move beyond simple replication and synthesize novel, complex chart code.

\paragraph{Iterative Self-Enhancement Loop.}
\label{ssec:self_boost}
Despite its quality, $\mathcal{C}_{cold}$ is typically scarce, and the coder trained on it exhibited limited structural complexity, making large-scale high-difficulty synthesis non-trivial. To address this, we introduce a self-enhancement mechanism that iteratively expands data and strengthens the coder capability.

Specifically, we first employ the cold-start coder to perform large-scale sampling at a temperature of 1.0, generating a raw candidate pool $\mathcal{C}_{raw}$ of 2M code samples. To ensure quality and diversity, We discard codes that trigger execution errors, then render the charts and filter out low-complexity samples using RPE. To ensure diversity, we exclude charts that exhibit high visual similarity to $\mathcal{I}_{hard}$ based on image embeddings from CLIP. We apply an RPE threshold $\geq 0.4$ and a similarity limit $\leq 0.65$ to retain only complex, non-redundant samples:
\begin{equation}
\label{eq:cboost}
\begin{split}
\mathcal{C}_{boost}=\big\{C\in&\mathcal{C}_{raw}\mid\;RPE(\mathcal{R}(C))\geq 0.4\\
&\land \mathrm{Sim}(\mathcal{R}(C),\mathcal{I}_{hard}) \leq 0.65\big\},
\end{split}
\end{equation}
where $\mathcal{R}(C)$ denotes the rendered image, and $\mathrm{Sim}(x,\mathcal{I}_{hard})=\max_{x'\in\mathcal{I}_{hard}}\cos(e(x),e(x'))$ calculates the maximum cosine similarity between the CLIP embedding $e(x)$ of a candidate and $\mathcal{I}_{hard}$.

Ultimately, we curate a boosted set $\mathcal{C}_{boost}$ of 200k samples and retrain the coder on $\mathcal{C}_{cold} \cup \mathcal{C}_{boost}$ to obtain a stronger model. We repeat this procedure for two iterations to ensure the model's generative stability and structural diversity.

\paragraph{Synthetic Chart Dataset.}
\label{ssec:dsyn}
We utilize the final complexity-aware coder to perform large-scale sampling at a temperature of 1.0, generating 1M raw chart code candidates $\mathcal{C}_{syn}$. We again apply RPE filtering to retain only high-complexity samples. This process yields a final dataset $\mathcal{D}_{syn}$ comprising 700k pairs of rendered images and their corresponding executable codes:
\begin{equation}
\label{eq:dsyn}
\begin{split}
\mathcal{D}_{syn}=\big\{(\mathcal{R}(C),&C)\mid\;C\in\mathcal{C}_{syn}\\
&\land RPE(\mathcal{R}(C))\ge 0.4 \big\}.
\end{split}
\end{equation}
This extensive corpus of complex charts serves as the foundation for our QA generation.

\subsection{Truth-Anchored Inverse QA Synthesis}
\label{ssec:qa_synthesis}
While the complexity of $\mathcal{D}_{\text{syn}}$ provides a foundation for constructing difficult questions, it introduces significant challenges in accurate answer extraction. To generate QA pairs that preserve both \textit{accuracy} and \textit{difficulty}, we propose to invert the generation flow to enforce "Answer as the Source of Truth".

\paragraph{Inverse Logic Construction.}
Existing QA generation approaches, which first synthesize a question and then employ a VLM to derive the answer ($Q \to A$), suffer from stochastic hallucinations and cannot guarantee correctness. To address this, we propose an Inverse Synthesis paradigm ($A \to Q$).

An advantage of $\mathcal{D}_{\text{syn}}$ is that the images are rendered from code, ensuring that codes encapsulate the absolute ground truth of all visual data. Leveraging this, we synthesize the answer first. As shown in Figure~\ref{fig:pipeline}, to ensure absolute precision, we employ Qwen3-30B-A3B-Thinking $\pi_{LLM}$ to analyze the chart code $C$ and generate a Python script $S$ that performs meaningful operations on the  data in chart code, yielding a precise numerical or categorical result. By executing this script in a deterministic environment $\mathcal{E}$, we obtain the ground-truth answer $A_{\text{py}}$, completely bypassing numerical errors from direct VLM answering:
{
\begin{equation}
    S \sim \pi_{\text{LLM}}(C), \quad A_{\text{py}} = \mathcal{E}(S).
\end{equation}
}

We then instruct the LLM again to reverse-engineer a question $Q$ that strictly aligns with chart code $C$ and Script $S$:
{
\begin{equation}
    Q \sim \pi_{\text{LLM}}(C, S).
\end{equation}
}

To guarantee that the synthesized question is unambiguously solvable, we perform a consistency check. We feed the question $Q$ and chart code $C$ back into the LLM to produce a solution $\hat{A}$, 
{
\begin{equation}
    \hat{A} \sim \pi_{\text{LLM}}(C, Q).
\end{equation}
}

\begin{table*}[t!]
\centering
\caption{Comparison of different models. The \colorbox{softblue}{shaded} rows denote our ChartVerse models.}
\small
\resizebox{1.0\textwidth}{!}{%
\begin{tabular}{lcccccccc}
\toprule
\textbf{Model} & \textbf{ChartQA} & \textbf{CharXiv} & \textbf{CharXiv} & \textbf{Chart} & \textbf{ChartX} &  \textbf{Evo} & \textbf{ChartBench} & \textbf{Avg}\\
&\textbf{Pro} & \textbf{(RQ)} & \textbf{(DQ)} & \textbf{Museum} & & \textbf{Chart} & \textbf{(GPT-acc)} & \\
\midrule
ECD-7B~\citep{yang2025effectivetrainingdatasynthesis} & 45.1 & 41.7 & 74.9 & 24.5 & 59.3 & 56.0 & 48.6 & 50.0\\
START-7B~\citep{liu2025startspatialtextuallearning} & 43.5 & 46.3 & 76.8 & 29.7 & 57.3 & 63.8 & 50.0 & 52.5\\
Chart-R1-7B~\citep{chen2025chartr1chainofthoughtsupervisionreinforcement} & 45.6 & 47.7 & 70.0 & 33.4 & 60.9 & 67.1 & 50.5 & 53.6\\
\midrule
\rowcolor{softblue} 
ChartVerse-2B & 48.2 & 46.9 & 71.2 & 37.5 & 60.5 & 66.8 & 49.1 & 54.3\\
InternVL3.5-38B~\citep{wang2025internvl35advancingopensourcemultimodal}  & 47.8 & 45.5 & 86.5 & 34.2 & 55.5 & 65.0 & 49.0 & 54.8\\
InternVL3.5-241B-A28B  & 50.7 & 47.5 & 88.1 & 36.1 & 59.1 & 67.4 & 48.6 & 56.8\\
Qwen3-VL-8B-Thinking~\cite{bai2025qwen3vltechnicalreport}  & 53.9 & 53.0 & 85.9 & 44.3 & 59.6 & 74.1 & 49.1 & 60.0\\
\rowcolor{softblue}
ChartVerse-4B & 55.2 & 56.2 & 84.1 & 45.9 & 63.7 & 75.0 & 52.9 & 61.9 \\
Qwen3-VL-30B-A3B-Thinking  & 55.8 & 56.6 & 86.9 & 49.2 & 62.3 & 77.2 & 52.4 & 62.9\\
\rowcolor{softblue}
ChartVerse-8B & 56.2 & 60.8 & 88.0 & 49.2 & 63.9 & 76.2 & 54.2 & 64.1\\
Qwen3-VL-32B-Thinking  & 58.8 & 65.2 & 90.2 & 55.9 & 64.1 & 80.8 & 54.3 & 67.0\\
Qwen3-VL-235B-A30B-Thinking & 60.0 & 66.1 & 90.5 & 60.0 & 64.5 & 79.9 & 54.5 & 67.9\\
\bottomrule
\end{tabular}
}
\label{tab:overall_results}
\end{table*}

We strictly retain samples where the inferred solution matches the programmatic ground truth:
\begin{equation}
    \mathcal{D}_{\text{pair}} = \{ (\mathcal{R}(C), Q, A_{\text{py}}) \mid \hat{A}=A_{\text{py}} \}.
\end{equation}

Through this rigorous inverse pipeline, we ensure that every sample is logically sound and factually accurate. It is worth noting that this logical construction phase operates exclusively within the textual domain. By relying solely on the code-text modality, we achieve high-efficiency synthesis while eliminating potential visual encoding errors.

\paragraph{CoT Distillation \& Difficulty Filtration.}
Correctness alone does not imply training value. We further estimate visual reasoning difficulty via repeated VLM rollouts and keep only samples that are neither trivial nor impossible. We employ Qwen3-VL-30B-A3B-Thinking $\pi_{VLM}$ to generate 3 CoT reasoning traces $\{\tau_1, \tau_2, \tau_3\}$.

As shown in Figure~\ref{fig:pipeline}, we extract the predicted answer $\hat{a}_j$ from each trace $\tau_j$ and calculate the failure rate $r(Q)$ against the ground truth:
{
\begin{equation}
    r(Q) = 1 - \frac{1}{3} \sum_{j=1}^{3} \text{Match}(\hat{a}_j, A_{\text{py}}).
\end{equation} 
}

We construct $\mathcal{D}_{\text{hard}}$ by retaining samples where $r(Q)$ is neither 0 nor 1. This ensures that samples are neither impossible to solve nor trivially simple.
\begin{equation}
    \mathcal{D}_{\text{hard}} = \{ (\mathcal{R}(C), Q, A_{\text{py}}, \tau) \mid 0 < r(Q) < 1 \}.
\end{equation}

We prioritize the most challenging samples with the highest $r(Q)$ in $\mathcal{D}_{\text{hard}}$ for the Reinforcement Learning dataset $\mathcal{D}_{\text{rl}}$, 
while utilizing the remaining samples enriched with correct CoT traces $\tau$ for the Supervised Finetuning dataset $\mathcal{D}_{\text{sft}}$. After the processing, we finally obtain the ChartVerse-SFT-600K and ChartVerse-RL-40K datasets. 
Details and prompts on synthesis pipelines and coder training are available in Appendix~\ref{app:data_synthesis}.

\begin{figure*}[t!]
     \centering
     \begin{subfigure}[b]{0.49\linewidth}
         \centering
         \includegraphics[width=0.95\linewidth]{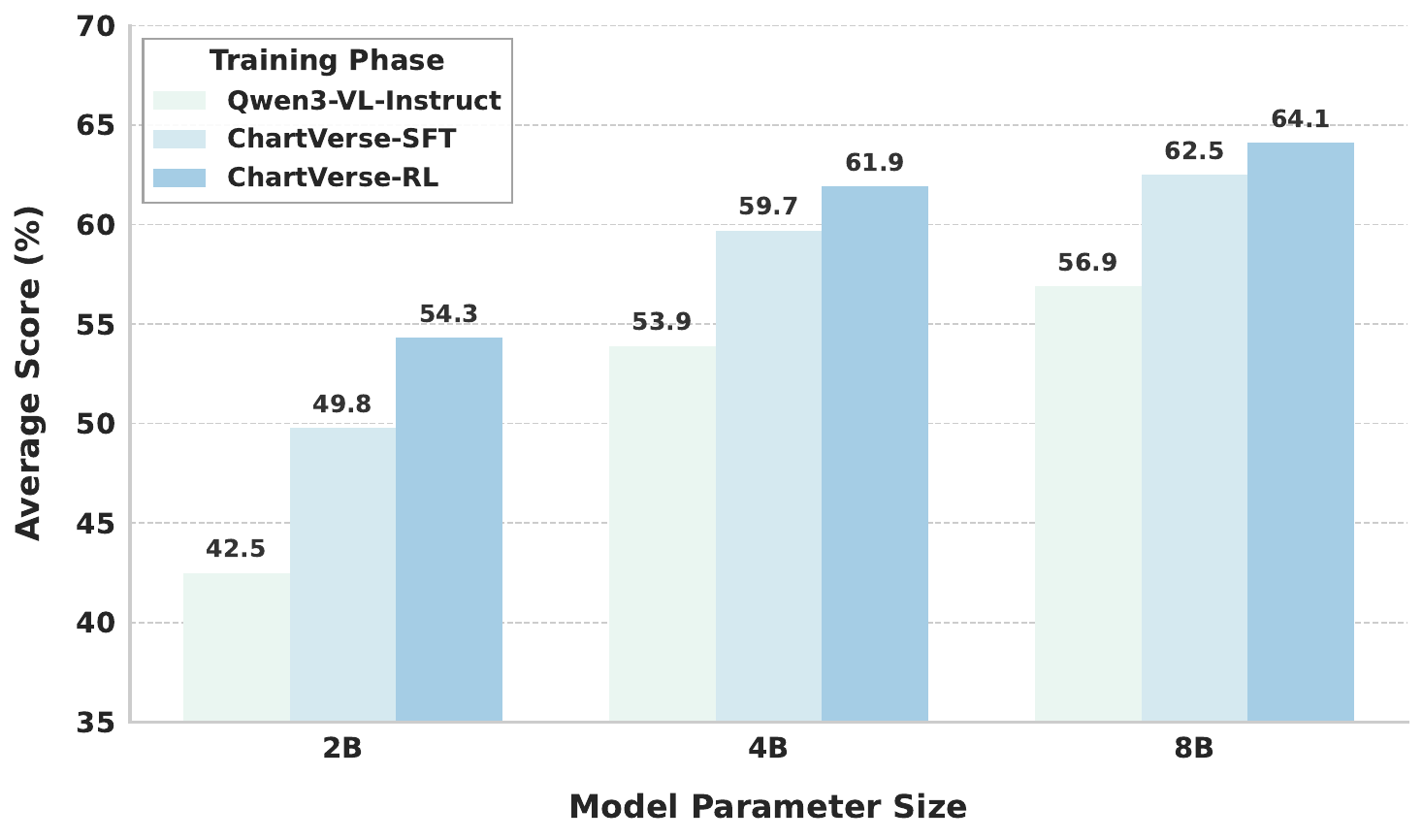}
         \caption{Performance on chart-related benchmarks}
         \label{fig:sft_rl_left}
     \end{subfigure}
     \hfill 
     \begin{subfigure}[b]{0.49\linewidth}
         \centering
         \includegraphics[width=0.95\linewidth]{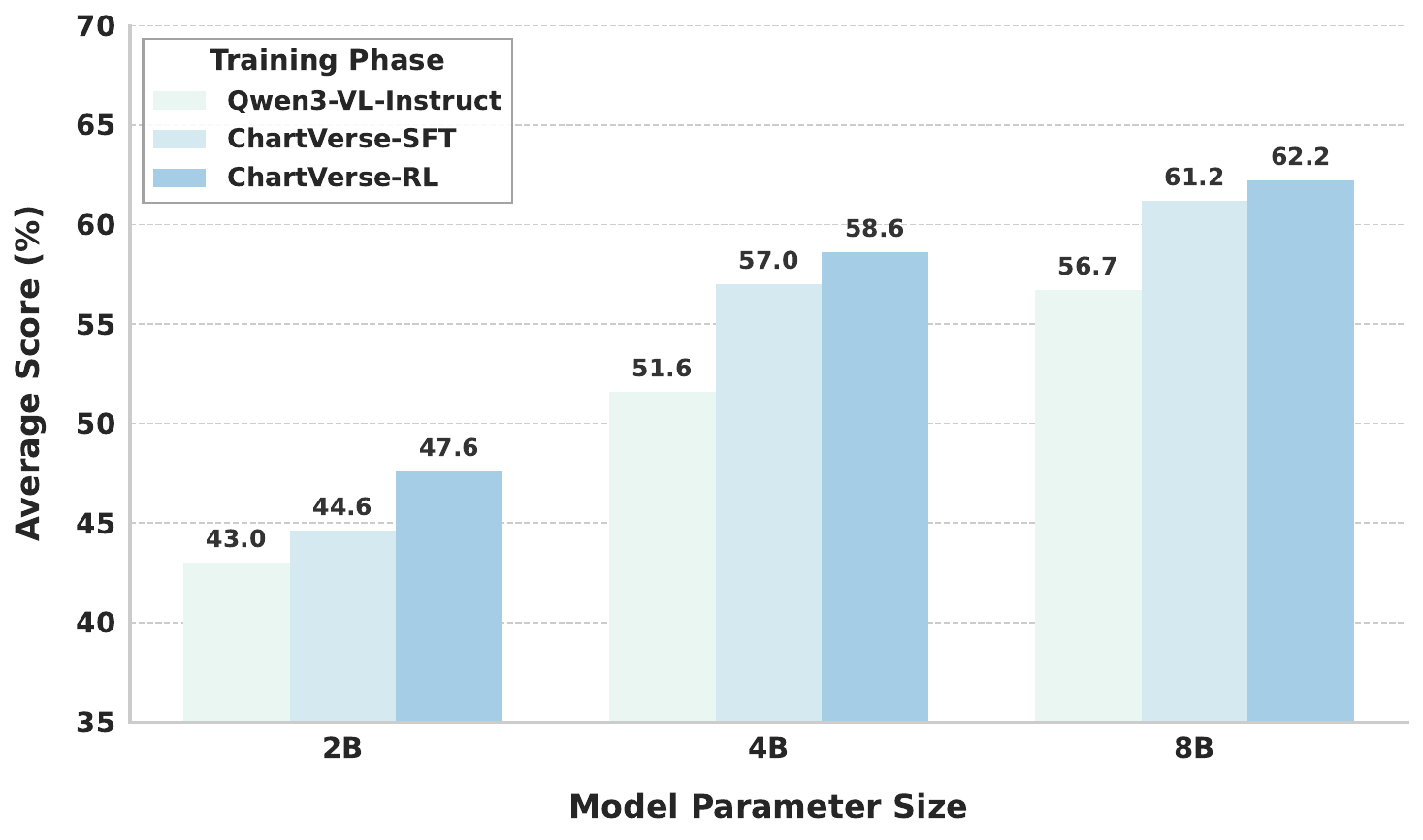}
         \caption{Performance on STEM-related benchmarks}
         \label{fig:stem_right}
     \end{subfigure}
     \caption{Performance evolution of ChartVerse across different training phases on chart and STEM benchmarks.}
     \label{fig:total_comparison}
\end{figure*}

\begin{figure*}[t!]
     \centering
     \begin{subfigure}[b]{0.49\linewidth}
         \centering
         \includegraphics[width=0.95\linewidth]{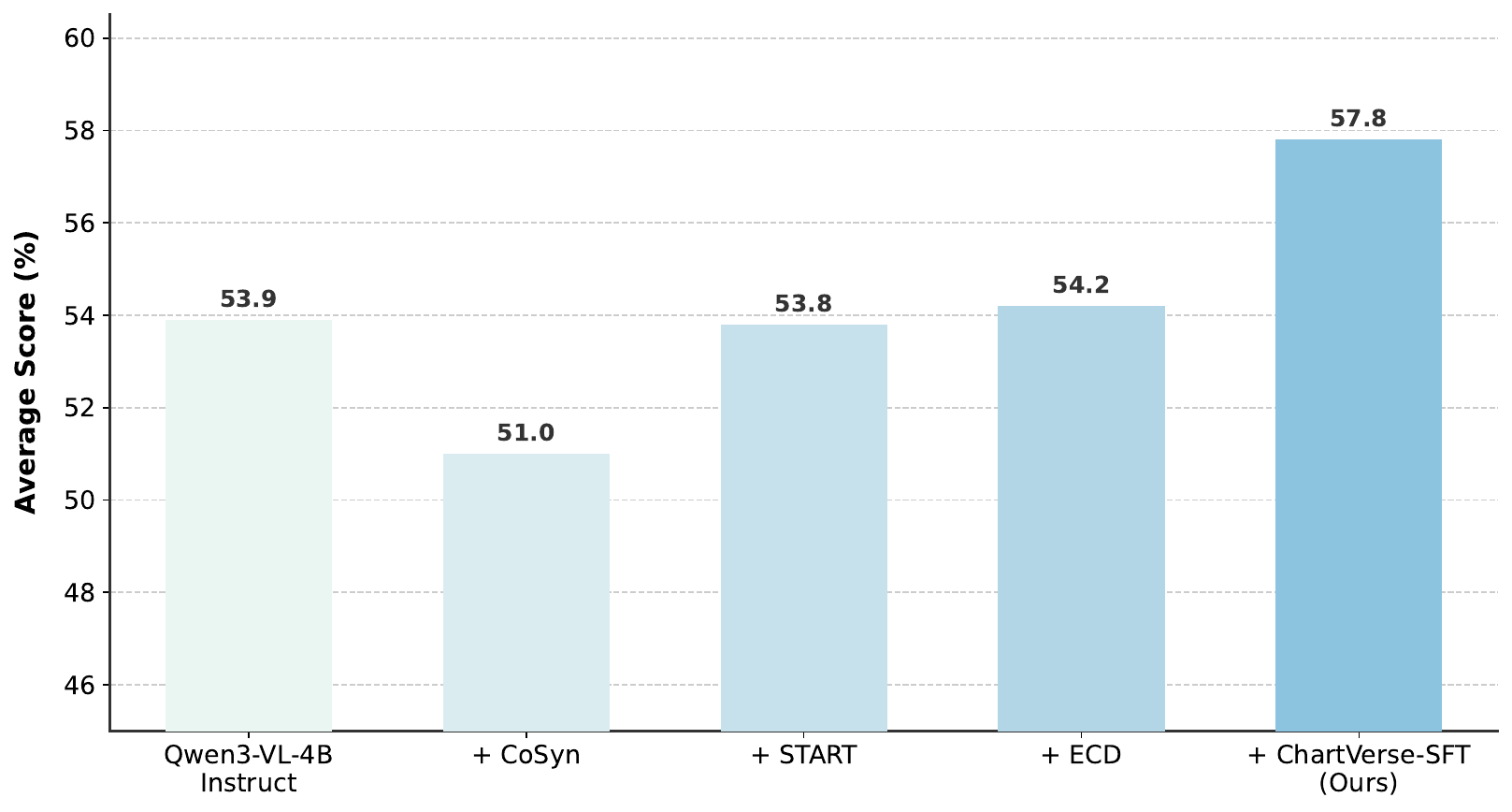}
         \caption{Comparison of different datasets}
         \label{fig:ablation_dataset}
     \end{subfigure}
     \hfill 
     \begin{subfigure}[b]{0.49\linewidth}
         \centering
         \includegraphics[width=0.95\linewidth]{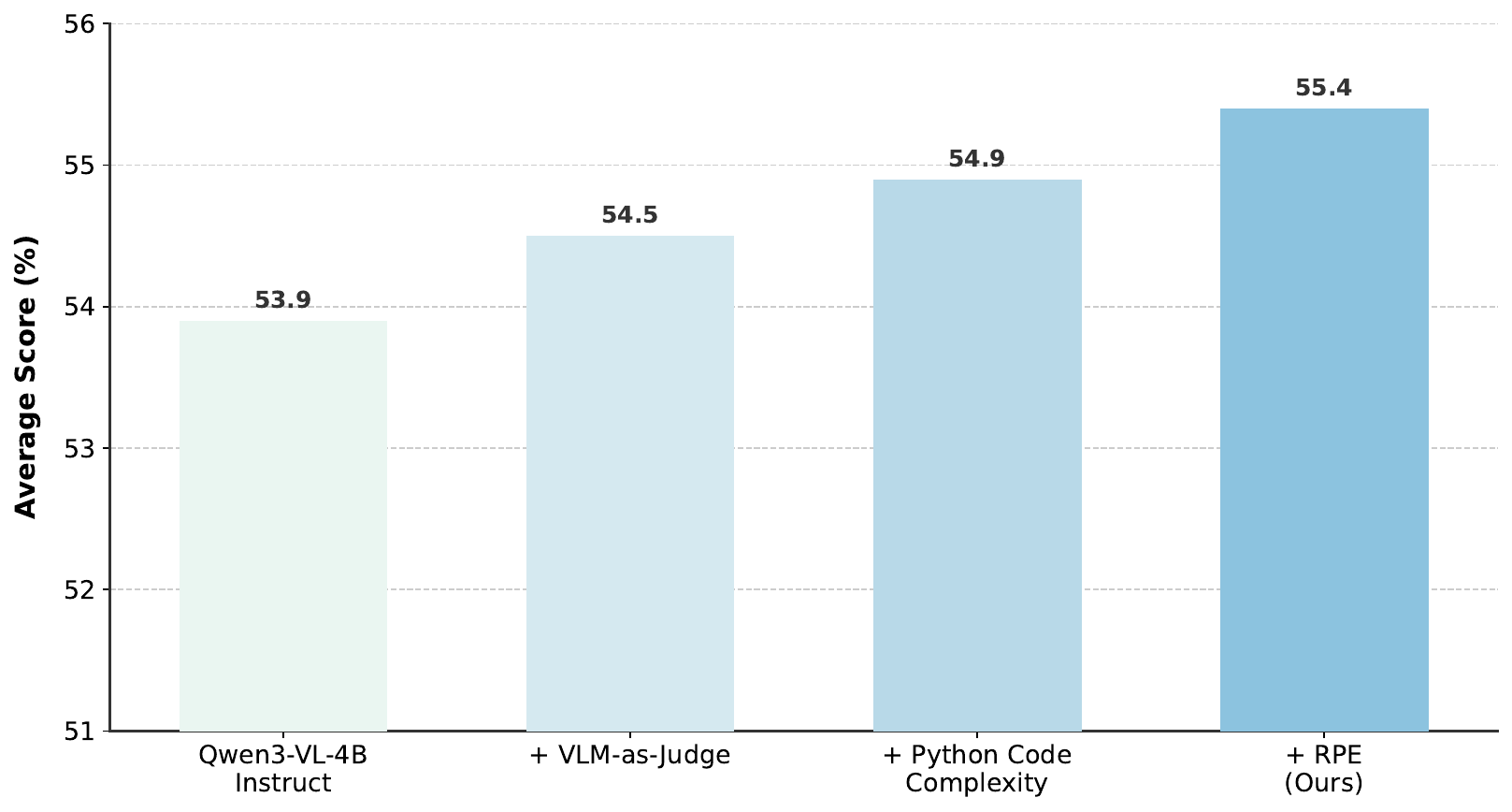}
         \caption{Ablation study on RPE strategy.}
         \label{fig:ablation_RPE}
     \end{subfigure}
     \caption{Ablation study results on different datasets and the proposed RPE strategy.}
     \label{fig:ablation_dataset_rpe}
\end{figure*}

\section{Experiments}
\label{Experiments}

\subsection{Experimental Setup}
\label{ssec:experimental_setup}





\paragraph{Training Details.}
All ChartVerse models are initialized from the Qwen3-VL-Instruct~\cite{bai2025qwen3vltechnicalreport} series. We first perform Supervised Fine-Tuning (SFT) on the ChartVerse-SFT-600K dataset using \texttt{LLaMA-Factory}~\citep{zheng2024llamafactoryunifiedefficientfinetuning}. Subsequently, Reinforcement Learning (RL) is conducted on the ChartVerse-RL-40K dataset with \texttt{VeRL}~\citep{Sheng_2025}, adopting the GSPO~\citep{zheng2025groupsequencepolicyoptimization} algorithm.
Comprehensive training configurations and hyperparameters are provided in Appendix~\ref{app:training_details}.

\paragraph{Baselines.}
We compare our tuned ChartVerse models against two categories of models:
(1) General open-source VLMs: leading models including Qwen3-VL-Thinking series (8B, 30B, 32B, 235B)~\citep{bai2025qwen3vltechnicalreport}, InternVL3.5 series (38B, 241B)~\citep{wang2025internvl35advancingopensourcemultimodal}. 
(2) Specialized chart-domain VLMs: ECD~\citep{yang2025effectivetrainingdatasynthesis},START~\citep{liu2025startspatialtextuallearning}, and Chart-R1~\citep{chen2025chartr1chainofthoughtsupervisionreinforcement}.

\paragraph{Benchmarks.}
We evaluate performance across 6 benchmarks necessitating complex chart understanding and reasoning: ChartQA-Pro~\citep{masry2025chartqaprodiversechallengingbenchmark}, CharXiv~\citep{wang2024charxivchartinggapsrealistic}, ChartMuseum~\citep{tang2025chartmuseumtestingvisualreasoning}, ChartX~\cite{xia2025chartxchartvlmversatile}, ChartBench~\citep{xu2024chartbenchbenchmarkcomplexvisual}, and EvoChart~\citep{huang2025evochartbenchmarkselftrainingapproach}.

\subsection{Main Results}
\paragraph{Overall Results.}
Table~\ref{tab:overall_results} compares ChartVerse models with above baselines. Overall, ChartVerse consistently delivers strong performance across all model scales, demonstrating the effectiveness of our difficulty-aware data synthesis framework. We highlight three key observations:

\noindent\textbf{(1) Competitive performance at small scale.}
ChartVerse-2B achieves an average score of 54.3, exceeding all chart-specific baselines, including ECD-7B, START-7B, and Chart-R1-7B. This shows that complexity-controlled chart data can substantially offset model size limitations.

\noindent\textbf{(2) Data quality over model scale.}
ChartVerse-4B attains an average score of 61.9, outperforming Qwen3-VL-8B-Thinking (60.0) despite using only half the parameters. This gap highlights the dominant role of data quality in improving chart reasoning performance.

\noindent\textbf{(3) Beyond the teacher model.}
ChartVerse-8B further improves the average score to 64.1, surpassing its teacher Qwen3-VL-30B-A3B-Thinking (62.9) and approaching Qwen3-VL-32B-Thinking (67.0). This result indicates that ChartVerse enables student models to exceed the distillation ceiling.

Collectively, these observations validate the advantages of ChartVerse’s rigorously controll chart and QA synthesis pipeline, showing that scalable generation of training data translates into consistent gains across model scales.


\subsection{Training Stages and Generalization}

\textbf{SFT and RL are Both Effective.}
Figure~\ref{fig:sft_rl_left} reports results on chart-related benchmarks under different training stages. Across all model sizes, ChartVerse-SFT yields substantial improvements over the Qwen3-VL-Instruct baseline, and ChartVerse-RL further provides consistent gains. For instance, ChartVerse-2B improves from 42.5 to 49.8 after SFT and reaches 54.3 after RL, while ChartVerse-8B increases from 56.9 to 62.5 and 64.1, respectively. These results indicate that SFT establishes strong chart reasoning foundations, and RL further enhances performance by focusing on more challenging samples.

\textbf{Strong Generalization to STEM tasks.}
We further evaluate ChartVerse on STEM-related benchmarks, including MathVista~\citep{mathvista}, DynaMath~\citep{dynamath}, MathVerse~\citep{mathverse}, LogicVista~\citep{logicvista}, and VisuLogic~\citep{visulogic}. As shown in Figure~\ref{fig:stem_right}, ChartVerse-trained models outperform the Instruct baseline across all scales. Notably, ChartVerse-8B improves from 56.7 to 61.2 after SFT and reaches 62.2 with RL, demonstrating that the reasoning skills learned from ChartVerse data transfer effectively to out-of-domain STEM reasoning tasks.
\begin{figure*}[!t]
\centering
\begin{subfigure}{0.47\textwidth}
    \centering
    \includegraphics[width=1.0\textwidth]{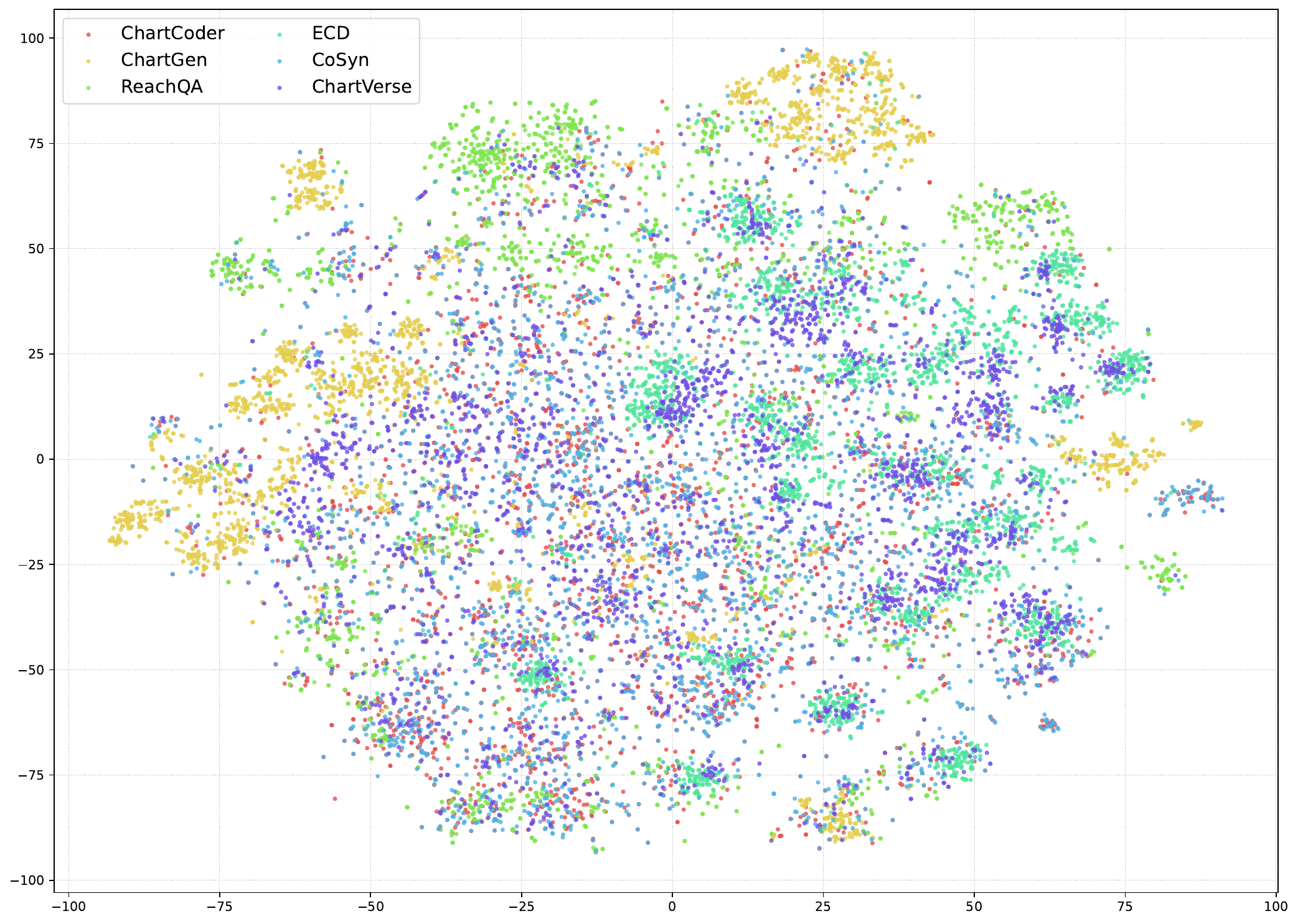} 
    \caption{Visualization of chart image embeddings}
    \label{fig:tsne_text_pretrain}
\end{subfigure}
\hfill
\begin{subfigure}{0.47\textwidth}
    \centering
    \includegraphics[width=1.0\textwidth]{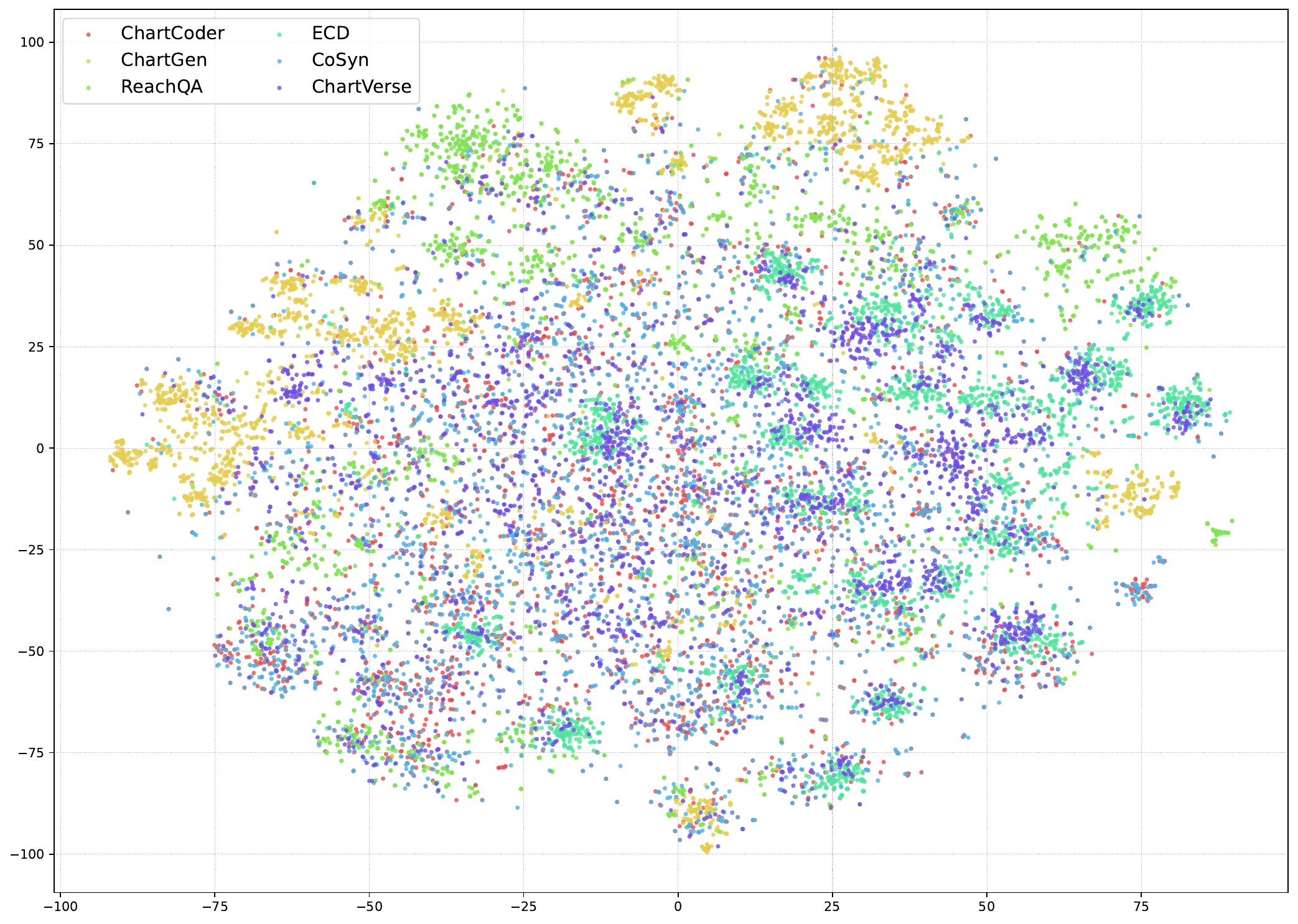}
    \caption{Visualization of chart code embeddings}
    \label{fig:tsne_image_pretrain}
\end{subfigure}
\caption{t-SNE visualizations of feature distributions for ChartVerse-SFT and existing datasets.}
\label{fig:tsne}
\end{figure*}

\begin{table*}[t!]
\centering
\caption{Performance comparison of different QA synthesis strategies. All models are trained on Qwen3-VL-4B-Instruct with 100k synthesized samples.}
\small
\resizebox{0.99\textwidth}{!}{%
\begin{tabular}{lcccccccc}
\toprule
\textbf{Training Data (100k)} & \textbf{ChartQA} & \textbf{CharXiv} & \textbf{CharXiv} & \textbf{Chart} & \textbf{ChartX} &  \textbf{Evo} & \textbf{ChartBench} & \textbf{Avg}\\
&\textbf{Pro} & \textbf{(RQ)} & \textbf{(DQ)} & \textbf{Museum} & & \textbf{Chart} & \textbf{(GPT-acc)} & \\
\midrule
Qwen3-VL-4B-Instruct & 53.7 & 39.7 & 76.2 & 37.2 & 57.2 & 68.2 & 45.1 & 53.9\\
\midrule
+ Image-Space Gen. & 53.1 & 46.6 & 81.3 & 38.4 & 58.4 & 70.4 & 47.3  & 56.5\\
+ Code-Space Gen. & 53.9 & 47.0 & 81.2 & 38.6 & 58.9 & 70.3 & 47.8 & 56.8 \\
+ Truth-Anchored Gen. w/o Failrate & 54.0 & 47.4 & 81.7 & 39.8 & 59.7 & 70.7 & 48.2 & 57.4 \\
\rowcolor{softblue}  + Truth-Anchored Gen. w/ Failrate (Ours) & 53.4 & 48.8 & 82.6 & 40.3 & 60.1 & 70.9 & 48.7 & 57.8\\
\bottomrule
\end{tabular}
}
\label{tab:truth-grounded_cmp}
\end{table*}
\section{Ablations and Analysis}
\label{sec:ablations}

\subsection{Fair Comparison with Existing Datasets}
To control for data scale, we compare different training datasets under a fixed budget of 100K samples. As summarized by the \emph{Avg} scores in Table~\ref{fig:ablation_dataset}, most existing datasets provide little benefit to Qwen3-VL-4B-Instruct: CoSyn even degrades performance (53.9 → 51.0), while START and ECD yield only marginal gains (53.8 and 54.2, respectively), barely surpassing the baseline.
In contrast, ChartVerse-SFT leads to a substantial improvement, boosting the average score to 57.8 (+3.9). This clear gap under identical data budgets indicates that the performance gains stem from data quality and task structure rather than scale, highlighting ChartVerse-SFT as significantly more effective for enhancing strong modern VLMs.

\subsection{Effectiveness of Rollout Posterior Entropy}
To verify the effectiveness of RPE, we compare RPE with two representative data selection strategies: \emph{VLM-as-Judge}, which selects samples with high model-assessed visual complexity, and \emph{Python Code Complexity}, which favors charts with dense Matplotlib structures. All strategies select 100K samples from the same data pool to ensure a fair comparison.

As shown in Figure~\ref{fig:ablation_RPE}, RPE identifies substantially harder samples, as reflected by the highest failure rate of Qwen3-VL-4B-Instruct (27.6\%), compared to 21.1\% for VLM-as-Judge and 23.5\% for Python Code Complexity. This suggests that RPE goes beyond surface-level visual or structural heuristics and captures samples that are intrinsically more challenging for the model.
Crucially, this increased difficulty translates into better downstream performance. Fine-tuning on RPE-selected data achieves the highest average score (55.4), outperforming both competing strategies. In contrast, methods based on perceived complexity yield smaller gains despite selecting easier samples. These results indicate that RPE is more effective at selecting \emph{learning-relevant} hard examples, leading to stronger performance improvements under a fixed data budget.

\subsection{Diversity and Distribution Analysis}
To analyze the quality and diversity of ChartVerse-SFT, we compare it with representative chart datasets from multiple perspectives, including data scale, supervision signals, and distributional diversity. For feature-level analysis, we extract embeddings from chart code and images using Qwen3-8B-Embedding and CLIP, and visualize their distributions with t-SNE.

As illustrated in Figure~\ref{fig:tsne}, ChartVerse-SFT covers a substantially broader feature space than prior datasets, largely subsuming the distributions of existing chart corpora. This indicates that our dataset is not concentrated around a narrow set of visual or structural patterns, but instead spans diverse chart styles and semantics.

We further quantify diversity using complementary metrics. Color Distribution Entropy measures the richness of visual appearance, while Semantic Embedding Spread captures structural and semantic variation across samples. As reported in Table~\ref{tab:chart_cmp}, ChartVerse-SFT consistently attains the highest values on both metrics, exceeding all existing datasets by a clear margin. Together with its higher rollout posterior entropy, these results suggest that ChartVerse-SFT is not only larger in scale, but also more diverse and information-rich, offering broader coverage of chart layouts, visual styles, and semantic structures.




\subsection{Breaking the Distillation Ceiling}

A central goal of ChartVerse is to enable the student model to outperform its teacher, which we achieve through truth-anchored inverse QA synthesis. We compare this method with two common QA construction strategies—direct image-based generation and direct code-based generation—and include an ablation that removes fail-rate based selection.

As shown in Table~\ref{tab:truth-grounded_cmp}, all QA synthesis strategies improve upon the baseline, indicating that synthetic supervision is generally beneficial. However, the magnitude of improvement varies substantially across methods. Direct image-space generation yields only moderate gains (Avg 56.5), while code-space generation performs slightly better (56.8), suggesting that structural cues from code are helpful but still limited.

In contrast, truth-anchored generation leads to a clear performance jump (57.4), demonstrating the advantage of anchoring QA synthesis to verifiable ground truth rather than surface-level patterns. Importantly, incorporating failure-rate-based selection further boosts the average score to 57.8, achieving the best performance. This additional gain highlights the role of hard-sample mining in maximizing the effectiveness of synthesized data, and confirms that our strategy produces not only more accurate but also more effective supervision.




\section{Conclusion}
We introduce ChartVerse, a scalable framework addressing the scarcity of complex data for chart reasoning. We introduce Rollout Posterior Entropy to quantify chart complexity, guiding Complexity-Aware Chart Coder to synthesize diverse, non-trivial charts. We propose Truth-Anchored Inverse QA Synthesis, ensuring accuracy via code-derived ground truths, reverse-synthesized questions, and consistency checks. Trained on our ChartVerse-SFT-600k and ChartVerse-RL-40k, ChartVerse-8B surpasses its teacher model Qwen3-VL-30B-A3B-Thinking and approaches Qwen3-VL-32B-Thinking, demonstrating the superior quality of our data. 

\section*{Limitations}

\textbf{Computational Overhead of RPE.} While Rollout Posterior Entropy effectively quantifies complexity, its reliance on large-scale sampling entails computational costs. Despite mitigation via vLLM, developing lightweight approximation methods to reduce resource demands remains a valuable direction for future optimization.

\textbf{Scope of Answer Types.} Our Truth-Anchored Inverse QA prioritizes rigorous correctness through code execution, resulting in datasets focused primarily on precise numerical and logical reasoning. Future work aims to extend this pipeline to include qualitative visual interpretations while preserving the same standards of verification.

\textbf{Teacher Model Capacity.} Our current synthesis employs Qwen3-VL-30B-A3B-Thinking. While effective, a capability gap remains compared to frontier models. Leveraging significantly larger models, such as Qwen3-VL-235B-A30B-Thinking, could further enhance the reasoning depth and complexity of the generated data.

\section*{Acknowledgments}
This work is supported by Shanghai Artificial Intelligence Laboratory, Fundamental and Interdisciplinary Disciplines Breakthrough Plan of the Ministry of Education of China (JYB2025XDXM113), National Natural Science Foundation of China (92470121, 62402016), National Key R\&D Program of China (2024YFA1014003), Zhongguancun Academy (C20250204, C20250602),  Beijing Major Science and Technology Project (Z251100008125043, Z251100008425023), and High-performance Computing Platform of Peking University.

\bibliography{custom}

\newpage
\clearpage
\appendix

\begin{center}
    \Large{\textbf{Appendix}}
\end{center}
\nolinenumbers

\section{More Training Details}
\label{app:training_details}

We implement our training pipeline using the LLaMA-Factory and Verl frameworks, conducted on a cluster of 4 nodes equipped with 8 A100 GPUs each (32 GPUs in total).

\paragraph{Supervised Fine-Tuning.}
We perform full-parameter fine-tuning on the \texttt{Qwen3-VL-Instruct} models. The visual encoder is configured to handle dynamic resolutions ranging from $802,816$ to $3,211,264$ pixels. To support extensive chart reasoning, we set a large context cutoff length of 22,000 tokens. To enhance training efficiency, we employ sequence packing and configure the training with a global batch size of 128. The model is trained for 1 epoch using the cosine learning rate scheduler with a peak learning rate of $1.0 \times 10^{-5}$ and a warmup ratio of 0.1. We utilize DeepSpeed ZeRO-3 optimization and BF16 to maximize memory efficiency during the full-parameter update. The entire fine-tuning process for the 8B model takes approximately 1.5 days.

\paragraph{Reinforcement Learning.}
Following SFT, we apply Group Relative Policy Optimization (GRPO) to further align the model's reasoning capabilities. We employ vLLM as the rollout engine, sampling $G=16$ distinct responses for each prompt with a temperature of $1.0$ and no top-$k$ restriction. The context window is expanded to support long-chain reasoning, with a maximum prompt length of 16,384 tokens and a maximum response length of 16,384 tokens. The actor model is optimized with a learning rate of $1.0 \times 10^{-6}$, utilizing a global batch size of 128. To ensure stability with such long contexts, we disable the standard KL-divergence penalty ($\beta_{KL}=0$) and instead rely on a conservative clipping strategy (ratio $\epsilon \in [3 \times 10^{-4}, 4 \times 10^{-4}]$). The RL process runs for approximately 300 steps. Training the 8B model in this stage requires approximately 4 days.

\section{Evaluation Details}
\label{app:evaluation_details}
We conduct a comprehensive evaluation using the VLMEvalKit~\citep{duan2025vlmevalkitopensourcetoolkitevaluating}. Since specific domain-specialized datasets such as ChartX, EvoChart, and ChartBench are not natively integrated, we preprocessed and converted these benchmarks into the standard VLMEvalKit interface format. For evaluation metrics, we replace traditional exact string matching with the compass-verifier~\citep{liu2025compassverifierunifiedrobustverifier}, which employs an LLM-as-a-Judge to accurately assess response correctness. Regarding rollout settings, following the guidelines from OpenDataArena~\citep{cai2025opendataarenafairopenarena,zhu2026factors} and the official Qwen documentation, we set the temperature to $0.6$, top-$p$ to $0.95$, and top-$k$ to $20$. Additionally, we apply a repetition penalty of $1.05$ and limit the maximum response length to $32768$ tokens. Furthermore, we exclude all system prompts during inference.

\section{Data Synthesis Details}
\label{app:data_synthesis}

\subsection{Complexity-Aware Chart Coder}

We first illustrate the distribution of our cold start dataset $\mathcal{C}_{\text{cold}}$, in Table~\ref{tab:cold_start_dist}. This diverse collection serves as the foundational corpus for initializing our coders. Furthermore, to generate high-quality code representations for complex charts, we utilize Claude-4-Sonnet. The specific prompt designed for this generation process is presented in Table~\ref{tab:claude_prompt}.

\begin{table}[h]
    \centering
    \caption{Data distribution of the cold start dataset $\mathcal{C}_{\text{cold}}$.}
    \label{tab:cold_start_dist}
    \begin{tabular}{lrr}
    \toprule
    \textbf{Source} & \textbf{Count} & \textbf{Percentage} \\
    \midrule
    CoSyn & 22088 & 35.67\% \\
    ChartGen & 15337 & 24.77\% \\
    ChartCoder & 10214 & 16.50\% \\
    ChartQA & 5602 & 9.05\% \\
    FigureQA & 4214 & 6.81\% \\
    PlotQA & 2150 & 3.47\% \\
    ECD & 1771 & 2.86\% \\
    ReachQA & 543 & 0.88\% \\
    \midrule
    \textbf{Total} & \textbf{61919} & \textbf{100.00\%} \\
    \bottomrule
    \end{tabular}
\end{table}

\begin{table*}[h!]
\centering
\caption{Performance evolution of ChartVerse across different training phases on chart and STEM benchmarks.}
\resizebox{\textwidth}{!}{%
\begin{tabular}{lccccccccccccccc}
\toprule
\multirow{2.5}{*}{\textbf{Model}} & \multicolumn{8}{c}{\textbf{Chart-Related}} & \multicolumn{5}{c}{\textbf{STEM-Related}} \\
\cmidrule(lr){2-9} \cmidrule(lr){10-15} & \textbf{ChartQA-Pro} & \textbf{CharXiv(DQ)} & \textbf{CharXiv(RQ)} & \textbf{ChartMuseum} & \textbf{ChartX} & \textbf{EvoChart} & \textbf{ChartBench} & \textbf{Avg} & \textbf{MathVista} & \textbf{DynaMath} & \textbf{MathVerse} & \textbf{LogicVista} & \textbf{VisuLogic} & \textbf{Avg} \\
\midrule
\hspace{3mm} Qwen3-VL-Instruct-2B & 42.1 & 26.8 & 62.3 & 23.9 & 49.8 & 53.6 & 38.8 & 42.5 & 61.3 & 52.1 & 54.2 & 35.8 & 11.5 & 43.0 \\
\hspace{3mm} + SFT & 44.4 & 40.8 & 69.1 & 30.0 & 56.9 & 61.0 & 46.4 & 49.8 & 58.7 & 47.6 & 55.8 & 43.2 & 17.5 & 44.6 \\
\hspace{3mm} + RL & 48.2 & 46.9 & 71.2 & 37.5 & 60.5 & 66.8 & 49.1 & 54.3 & 60.4 & 49.2 & 56.8 & 48.5 & 23.1 & 47.6 \\
\midrule
\hspace{3mm} Qwen3-VL-Instruct-4B & 53.7 & 39.7 & 76.2 & 37.2 & 57.2 & 68.2 & 45.1 & 53.9 & 73.7 & 46.8 & 65.3 & 53.2 & 19.0 & 51.6 \\
\hspace{3mm} + SFT & 52.9 & 52.8 & 84.5 & 42.5 & 61.9 & 73.3 & 50.3 & 59.7 & 70.9 & 60.7 & 71.1 & 57.3 & 25.0 & 57.0\\
\hspace{3mm} + RL & 55.2 & 56.2 & 84.1 & 45.9 & 63.7 & 75.0 & 52.9 & 61.9 & 72.8 & 62.0 & 71.8 & 59.3 & 27.0 & 58.6\\
\midrule
\hspace{3mm} Qwen3-VL-Instruct-8B & 54.4 & 46.4 & 83.0 & 39.6 & 58.2 & 70.2 & 46.6 & 56.9 & 77.2 & 62.1 & 67.7 & 55.3 & 22.5 & 56.7\\
\hspace{3mm} + SFT & 55.5 &56.2 & 88.3 & 47.5 & 61.0 & 76.7 & 52.2 & 62.5 & 75.0 & 67.4 & 75.6 & 61.3 & 26.5 & 61.2 \\
\hspace{3mm} + RL & 56.2 & 60.8 & 88.0 & 49.2 & 63.9 & 76.2 & 54.2 & 64.1 & 75.6 & 69.0 & 76.5 & 62.6 & 27.1 & 62.2\\
\bottomrule
\end{tabular}%
}
\label{tab:split_results}
\end{table*}

\begin{table*}[t!]
\centering
\caption{Comparison of Different Datasets. All models are trained on Qwen3-VL-4B-Instruct using 100K samples.}
\small
\resizebox{0.99\textwidth}{!}{%
\begin{tabular}{lcccccccc}
\toprule
\textbf{Training Data (100K)} &  \textbf{ChartQA} & \textbf{CharXiv} & \textbf{CharXiv} & \textbf{Chart} & \textbf{ChartX} &  \textbf{Evo} & \textbf{ChartBench} & \textbf{Avg}\\
&\textbf{Pro} & \textbf{(RQ)} & \textbf{(DQ)} & \textbf{Museum} & & \textbf{Chart} & \textbf{(GPT-acc)} & \\
\midrule
Qwen3-VL-4B-Instruct & 53.7 & 39.7 & 76.2 & 37.2 & 57.2 & 68.2 & 45.1 & 53.9\\
\midrule
+ CoSyn & 50.3 & 35.1 & 74.4 & 33.5 & 55.8 & 62.4 & 45.8 & 51.0\\
+ START & 53.5 & 39.2 & 77.2 & 37.9 & 56.4 & 68.0 & 44.5 & 53.8\\
+ ECD & 54.4 & 38.3 & 77.5 & 37.7 & 58.3 & 68.8 & 44.7 & 54.2 \\
\rowcolor{softblue} + ChartVerse-SFT (Ours) & 53.4 & 48.8 & 82.6 & 40.3 & 60.1 & 70.9 & 48.7 & 57.8 \\
\bottomrule
\end{tabular}
}
\label{tab:fair_cmp}
\end{table*}

\begin{table*}[t!]
\centering
\caption{Comparison of data selection strategies. The Fail Rate (left) indicates sample difficulty, while the right section shows the results of Qwen3-VL-4B-Instruct trained on 100k selected data.}
\small
\resizebox{1.0\textwidth}{!}{%
\begin{tabular}{lc|cccccccc}
\toprule
& \multicolumn{1}{c|}{\textbf{Difficulty}} & \multicolumn{8}{c}{\textbf{Downstream Performance (\%)}} \\
\cmidrule(lr){2-2} \cmidrule(l){3-10}
\textbf{Strategy} & \textbf{Fail Rate}  & \textbf{ChartQA} & \textbf{CharXiv} & \textbf{CharXiv} & \textbf{Chart} & \textbf{ChartX} & \textbf{Evo} & \textbf{ChartBench} & \textbf{Avg}\\
& \textbf{(N=500)} & \textbf{Pro} & \textbf{(RQ)} & \textbf{(DQ)} & \textbf{Museum} & & \textbf{Chart} & \textbf{(GPT-acc)} \\
\midrule
Qwen3-VL-4B-Instruct & - & 53.7 & 39.7 & 76.2 & 37.2 & 57.2 & 68.2 & 45.1 & 53.9\\
\midrule
+ VLM-as-Judge & 21.1\% & 53.4 & 40.3 & 75.7 & 38.8 & 58.6 & 69.2 & 45.4 & 54.5\\
+ Python Code Complexity & 23.5\% & 53.7 & 40.5 & 77.7 & 39.0 & 58.5 & 70.3 & 44.7 & 54.9 \\
\rowcolor{softblue} + RPE (Ours) & 27.6\%  & 53.3 & 41.0 & 78.6 & 39.2 & 58.5 & 70.8 & 46.3 & 55.4\\
\bottomrule
\end{tabular}
}
\label{tab:rpe_cmp}
\end{table*}

\paragraph{Training Configuration.}
We train the Complexity-Aware Chart Coder using \texttt{Qwen2.5-Coder-7B-Instruct} as the backbone model. During both training and inference, we employ the identical system prompt $\mathcal{T}$, as presented in Table~\ref{tab:system_coder_prompt}. The training pipeline is implemented on the LLaMA-Factory framework. We apply the same training configuration across every iterative training stage to ensure consistency. We perform full-parameter fine-tuning with a global batch size of 16 and a context cutoff length of 4,096 tokens, utilizing sequence packing to maximize computational efficiency. The optimization process runs for 5 epochs using a cosine learning rate scheduler, configured with a peak learning rate of $2.0 \times 10^{-5}$ and a warmup ratio of 0.05. We employ DeepSpeed ZeRO-3 optimization with BF16 precision.

\paragraph{Sampling Configuration.}
During the large-scale sampling phase, we set the sampling temperature to $1.0$, top-$p$ to $0.95$, and top-$k$ to $20$, ensuring diverse and robust generation. Regarding the RPE, we processed a total of approximately 4 million samples. Leveraging vLLM, our RPE computation is highly efficient, completed in 3 days on 64 A100 GPUs.

\begin{table*}[t!]
\centering
\caption{Comparison of ChartVerse performance across Instruct and Thinking model bases.}
\small
\resizebox{0.99\textwidth}{!}{%
\begin{tabular}{lcccccccc}
\toprule
\textbf{Model Base} & \textbf{ChartQA} & \textbf{CharXiv} & \textbf{CharXiv} & \textbf{Chart} & \textbf{ChartX} & \textbf{Evo} & \textbf{Chart} & \textbf{Avg} \\
& \textbf{Pro} & \textbf{(RQ)} & \textbf{(DQ)} & \textbf{Museum} & & \textbf{Chart} & \textbf{Bench} & \\
\midrule
Qwen3-VL-4B-Instruct & 55.2 & 56.2 & 84.1 & 45.9 & 63.7 & 75.0 & 52.9 & 61.9 \\
Qwen3-VL-4B-Thinking & 55.8 & 57.6 & 85.0 & 47.7 & 64.1 & 76.1 & 54.5 & 63.0 \\
\midrule
Qwen3-VL-8B-Instruct & 56.2 & 60.8 & 88.0 & 49.2 & 63.9 & 76.2 & 54.2 & 64.1 \\
Qwen3-VL-8B-Thinking & 57.3 & 61.6 & 88.2 & 51.4 & 65.2 & 77.0 & 55.1 & 65.1 \\
\bottomrule
\end{tabular}
}
\label{tab:base_ablation}
\end{table*}

\begin{table*}[t!]
\centering
\caption{Ablation of teacher model scale. All results use Qwen3-VL-8B-Instruct as the student model.}
\small
\resizebox{0.99\textwidth}{!}{%
\begin{tabular}{llcccccccc}
\toprule
\textbf{Training Data} & \textbf{Teacher Model} & \textbf{ChartQA} & \textbf{CharXiv} & \textbf{CharXiv} & \textbf{Chart} & \textbf{ChartX} & \textbf{Evo} & \textbf{Chart} & \textbf{Avg} \\
& & \textbf{Pro} & \textbf{(RQ)} & \textbf{(DQ)} & \textbf{Museum} & & \textbf{Chart} & \textbf{Bench} & \\
\midrule
ChartVerse-SFT-600K & 30B-A3B-Thinking & 55.5 & 56.2 & 88.3 & 47.5 & 61.0 & 76.7 & 52.2 & 62.5 \\
ChartVerse-SFT-100K & 235B-A30B-Thinking & 56.9 & 59.2 & 88.7 & 49.9 & 63.4 & 76.5 & 54.0 & 64.1 \\
\bottomrule
\end{tabular}
}
\label{tab:teacher_ablation}
\end{table*}

\paragraph{RPE Data Generation.}
Regarding the RPE, we processed a total of approximately 4 million samples. This process involved approximately 32 million inference calls to the \texttt{Qwen3-VL-2B-Thinking} model. Leveraging vLLM~\cite{kwon2023efficientmemorymanagementlarge}, our RPE computation is highly efficient, completed in 3 days on 64 A100 GPUs

\paragraph{Data Decontamination.} 
To ensure zero overlap between training and evaluation data, we perform rigorous decontamination on the initial image pool $\mathcal{I}_{pool}$, ChartVerse-SFT-600K, and ChartVerse-RL-40K. Specifically, we discard any training sample whose chart image exhibits a CLIP cosine similarity exceeding 0.65 with any benchmark image. Notably, because our training charts are synthesized from scratch via code generation rather than sourced from existing datasets, the number of filtered images is negligibly small.

\subsection{Truth-Anchored Inverse QA Synthesis}

\paragraph{Inverse Logic Construction.}
Our synthesis pipeline relies on a sequence of carefully designed prompts to ensure logical consistency and factual accuracy. We present the specific prompts utilized for Python script generation, reverse question synthesis, and consistency checking in Table~\ref{tab:script_prompt}, Table~\ref{tab:qa_prompt}, and Table~\ref{tab:consistency_prompt}, respectively.

Specifically, in the Python script generation phase (Table~\ref{tab:script_prompt}), we enforce a strict constraint: the generated code must execute to yield a precise numerical value or a definitive categorical label, ensuring the answer is deterministic. Following this, during the reverse question synthesis (Table~\ref{tab:qa_prompt}), we mandate that the synthesized text must be a valid interrogative sentence, and the answer to this question must correspond exactly to the execution result of the preceding Python script.

In our specific implementation, we synthesize two distinct Python scripts for each chart code, followed by separate reverse question synthesis and consistency checks. This process involved a total of 4 million calls to \texttt{Qwen-30B-A3B-Thinking}, taking approximately 4 days on 128 A100 GPUs.

\paragraph{CoT Distillation \& Difficulty Filtration.}
We display the prompt used for distilling Chain-of-Thought (CoT) reasoning traces in Table~\ref{tab:distill_prompt}. 

To ensure high-quality and non-redundant reasoning traces, we apply a multi-stage filtering procedure to the distilled outputs:

\begin{itemize}
    \item \textbf{Template and Length Validation}: We first impose strict structural validation to ensure the usability of the distilled output. Specifically, we filter out any reasoning trace that fails to adhere to the mandated \verb|<think>...</think>| and \verb|<answer>...</answer>| output template. Furthermore, to prevent the retention of superficial or trivial rationales, we enforce a minimum length constraint, discarding traces that are shorter than 100 words.
    \item \textbf{N-gram De-duplication}: We detect and remove templated or overly repetitive CoTs using an n-gram overlap criterion. Concretely, we flag CoTs that contain any 50-gram that repeats at least 3 times. Flagged traces are discarded.
\end{itemize}

In this step, we processed approximately 1 million samples. With each sample requiring three distinct inference calls, this amounted to a total of 3 million calls to \texttt{Qwen3-VL-30B-A3B-Thinking}, consuming approximately 5 days on 128 A100 GPUs.

Throughout all processes in this stage, we set the LLM inference parameters to a temperature of $0.6$, top-$p$ of $0.95$, top-$k$ of $20$, and a maximum output length of $32768$.

\begin{table*}[t!]
\centering
\caption{Comparison of different models for RPE calculation. Results are computed on a random subset of 10K charts. \textbf{Range} denotes the difference between the 75th and 25th percentiles.}
\small
\resizebox{0.8\textwidth}{!}{%
\begin{tabular}{lcccccc}
\toprule
\textbf{Model} & \textbf{Mean} & \textbf{Std} & \textbf{25th} & \textbf{75th} & \textbf{Effective} & \textbf{Execution} \\
 & \textbf{RPE} & \textbf{RPE} & \textbf{Perc.} & \textbf{Perc.} & \textbf{Range} & \textbf{Success Rate} \\
\midrule
Qwen3-VL-2B-Instruct & 0.48 & 0.16 & 0.38 & 0.55 & 0.17 & 63.2\% \\
\rowcolor{softblue} Qwen3-VL-2B-Thinking & 0.41 & 0.18 & 0.32 & 0.51 & 0.19 & 88.6\% \\
Qwen3-VL-4B-Thinking & 0.35 & 0.10 & 0.30 & 0.42 & 0.12 & 93.3\% \\
Qwen3-VL-8B-Thinking & 0.32 & 0.09 & 0.27 & 0.37 & 0.10 & 96.1\% \\
\bottomrule
\end{tabular}
}
\label{tab:rpe_ablation}
\end{table*}

\section{Additional Results}
\subsection{Detailed Performance Analysis of Training Phases}
We provide the specific performance corresponding to Figure~\ref{fig:total_comparison} in Table ~\ref{tab:split_results}. The results across the 2B, 4B, and 8B scales demonstrate a consistent upward trajectory. ChartVerse-8B achieves a significant leap on CharXiv (RQ), rising from a baseline of 46.4 to 60.8. Similarly, on ChartMuseum, it reaches a peak performance of 49.2. The reasoning skills acquired from chart-specific data also generalize effectively to out-of-domain STEM tasks. ChartVerse-8B improves its performance on MathVerse from 67.5 to 76.5 and on DynaMath from 62.1 to 69.0. These gains confirm that the logical reasoning capabilities learned through our framework transfer robustly to complex scientific contexts.

\subsection{Comparison with Existing Datasets}
Complementing the visual comparison in Figure~\ref{fig:ablation_dataset}, Table~\ref{tab:fair_cmp} evaluates ChartVerse-SFT against competing datasets under a fixed 100K sample budget. While datasets like CoSyn and START degrade performance on specific benchmarks, ChartVerse-SFT consistently drives improvements. Specifically, our dataset boosts ChartMuseum to 40.3 and significantly lifts CharXiv (RQ) performance from 39.7 to 48.8, confirming that our gains stem from superior data quality rather than scale.

\subsection{Ablation on RPE Strategy}
The effectiveness of the Rollout Posterior Entropy strategy, as shown in Figure~\ref{fig:ablation_RPE}, is further quantified in Table~\ref{tab:rpe_cmp}. RPE successfully identifies intrinsically harder samples, evidenced by a model failure rate of 27.6, compared to only 21.1 for VLM-as-Judge. This focus on difficulty translates directly to downstream accuracy: on the CharXiv (DQ) benchmark, the RPE-trained model achieves 78.6, outperforming both VLM-as-Judge 75.7 and Python Code Complexity 77.7.

\subsection{Impact of the Model Base}
We investigate the compatibility of ChartVerse data with different model architectures by comparing Instruct and Thinking base models. Our primary experiments employ the Instruct series to rigorously validate that the ChartVerse dataset can inject complex reasoning capabilities from scratch, without relying on built-in extended thinking pre-training. To evaluate generalizability, we apply the full training recipe (ChartVerse-SFT-600K + ChartVerse-RL-40K) to both Instruct and Thinking models at the 4B and 8B scales. 

As shown in Table \ref{tab:base_ablation}, the Thinking bases yield additional improvements over the Instruct bases, confirming that our synthesized data is highly compatible with and beneficial for thinking-capable models. However, the performance gap between the two bases is relatively narrow. We attribute this to the extensive scale of ChartVerse-SFT-600K, which substantially reshapes the Instruct model's reasoning behavior, endowing it with analytical capabilities comparable to the native Thinking variants.

\subsection{Impact of the Teacher Model Scale}
Our primary data synthesis pipeline utilizes Qwen3-VL-30B-A3B-Thinking as the teacher to balance data quality and computational feasibility. By leveraging truth-anchored synthesis and failure-rate filtering, our framework successfully enables the student model to surpass the teacher despite the teacher's relatively constrained capacity. 

To explore the effect of scaling the teacher model, we conduct an ablation study using a significantly larger teacher, Qwen3-VL-235B-A30B-Thinking, to synthesize a subset of 100K high-quality SFT samples. As presented in Table \ref{tab:teacher_ablation}, training the 8B Instruct model on this 100K dataset achieves an average score of 64.1, which already outperforms the model trained on the full 600K dataset generated by the 30B teacher. This confirms that a stronger teacher directly enhances the synthesized data quality, and validates that our ChartVerse framework scales effectively with teacher capacity, yielding superior student performance even with a reduced sample size.

\section{Model Selection for RPE Computation}
The Relative Prediction Entropy (RPE) serves as our core metric for assessing chart complexity. We select \textbf{Qwen3-VL-2B-Thinking} as the base model for RPE calculation with 8 inference passes ($K=8$), based on the following three considerations:

\paragraph{Computational Efficiency.} RPE requires multiple forward passes per image. Given our initial pool of over 3M images, the 2B-scale model offers the most practical trade-off, enabling large-scale entropy estimation within a reasonable computational budget.

\paragraph{Complexity Discrimination.} Smaller models exhibit higher sensitivity to structural complexity. As shown in Table \ref{tab:rpe_ablation}, larger models (4B, 8B) reconstruct complex charts more consistently, which compresses the RPE distribution (standard deviation decreases from 0.18 to 0.09). The 2B model provides the widest RPE range and the highest standard deviation, offering the sharpest discriminative power to separate simple and complex charts.

\paragraph{Execution Reliability.} While Instruct and Thinking variants at the 2B scale show similar entropy spreads, the Thinking variant achieves a significantly higher code execution success rate (88.6\% vs. 63.2\%). Since RPE relies on successfully rendered images to compute valid entropy, the structural reasoning of the Thinking model minimizes noise from failed executions, ensuring more stable and trustworthy complexity scores.

\section{ChartVerse-SFT Analysis}
\label{app:chartverse_sft}

\begin{figure*}[!t]
    \centering
    \includegraphics[width=16cm]{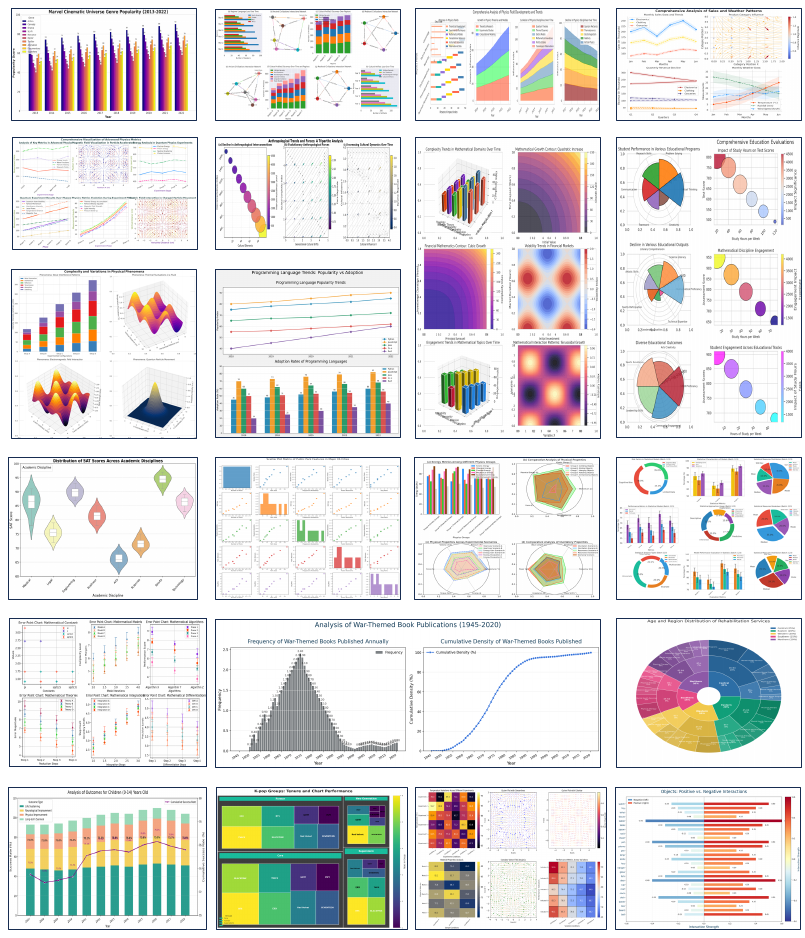}
    \caption{Overview of the ChartVerse-SFT dataset. The samples demonstrate high diversity in chart types including 3D, hierarchical, and statistical plots, as well as structural complexity featuring various multi-subplot layouts.}
    \label{fig:chartverse_examples}
\end{figure*}

We first demonstrate the visual complexity of our synthesized training corpus, ChartVerse-SFT. As illustrated in Figure~\ref{fig:chartverse_examples}, the dataset exhibits exceptional diversity by covering a wide spectrum of visualization types. These range from standard statistical graphs such as violin plots and radar charts to high-dimensional 3D representations and hierarchical structures like treemaps. Crucially, a significant portion of the data features intricate multi-subplot layouts and mixed-type dashboards. This structural complexity compels the model to handle diverse visual encoding schemas and perform fine-grained reasoning across spatially distinct panels, significantly surpassing the difficulty of traditional chart datasets.

\begin{figure*}[!t]
    \centering
    \includegraphics[width=16cm]{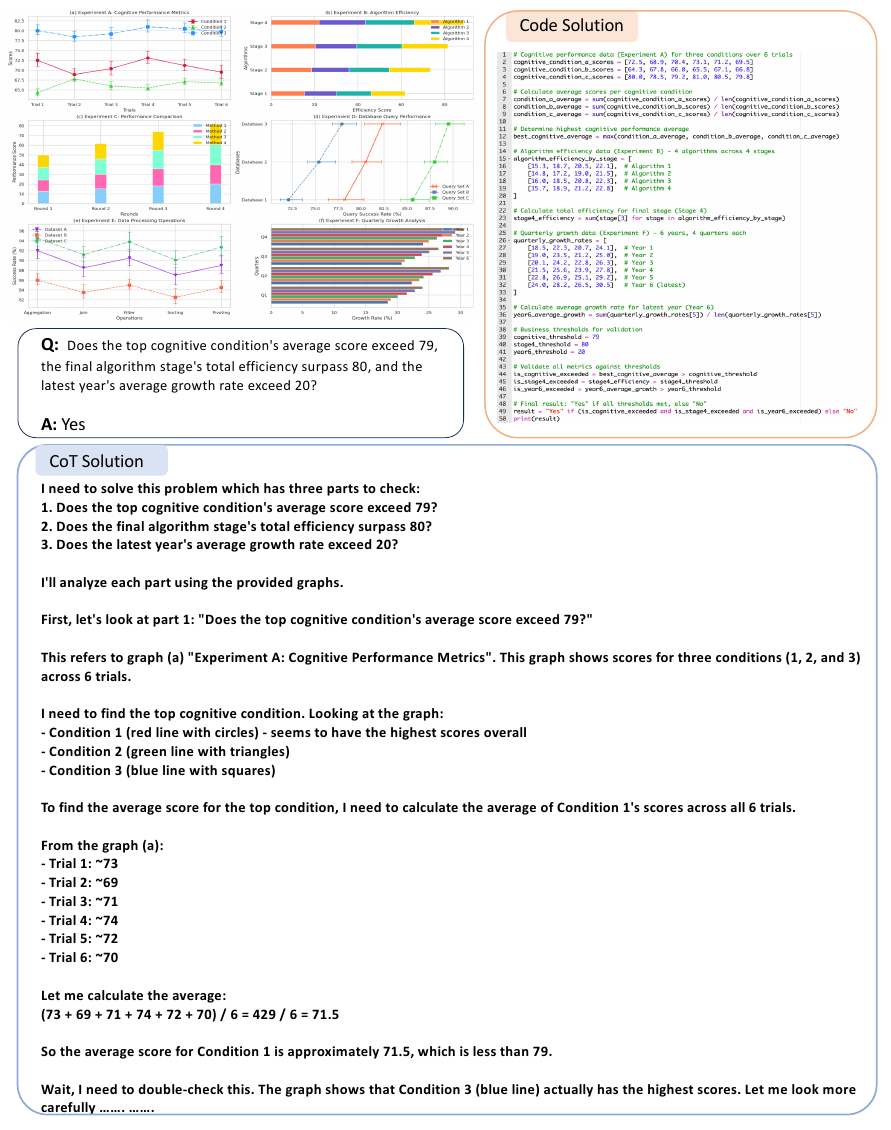}
    \caption{ChartVerse-SFT QA Example-1}
    \label{fig:qa_1}
\end{figure*}

\begin{figure*}[!t]
    \centering
    \includegraphics[width=16cm]{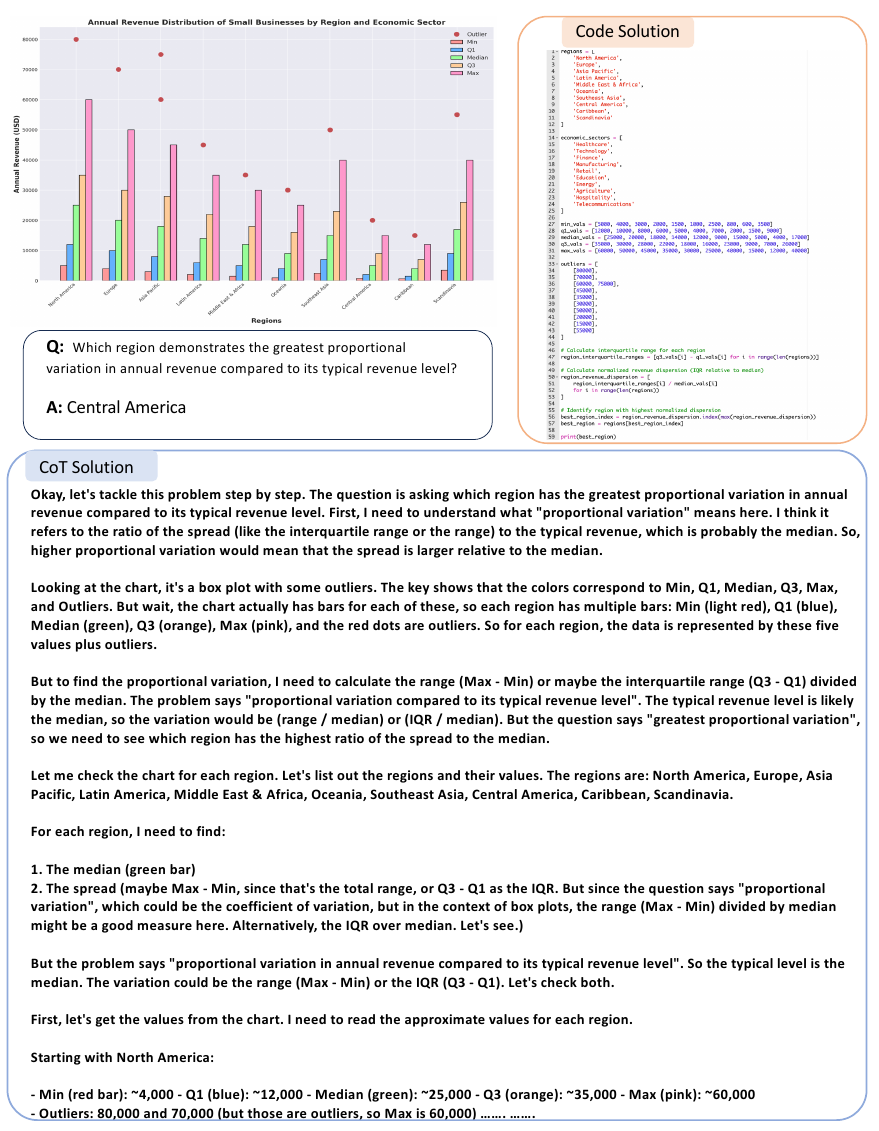}
    \caption{ChartVerse-SFT QA Example-2}
    \label{fig:qa_2}
\end{figure*}

\begin{figure*}[!t]
    \centering
    \includegraphics[width=16cm]{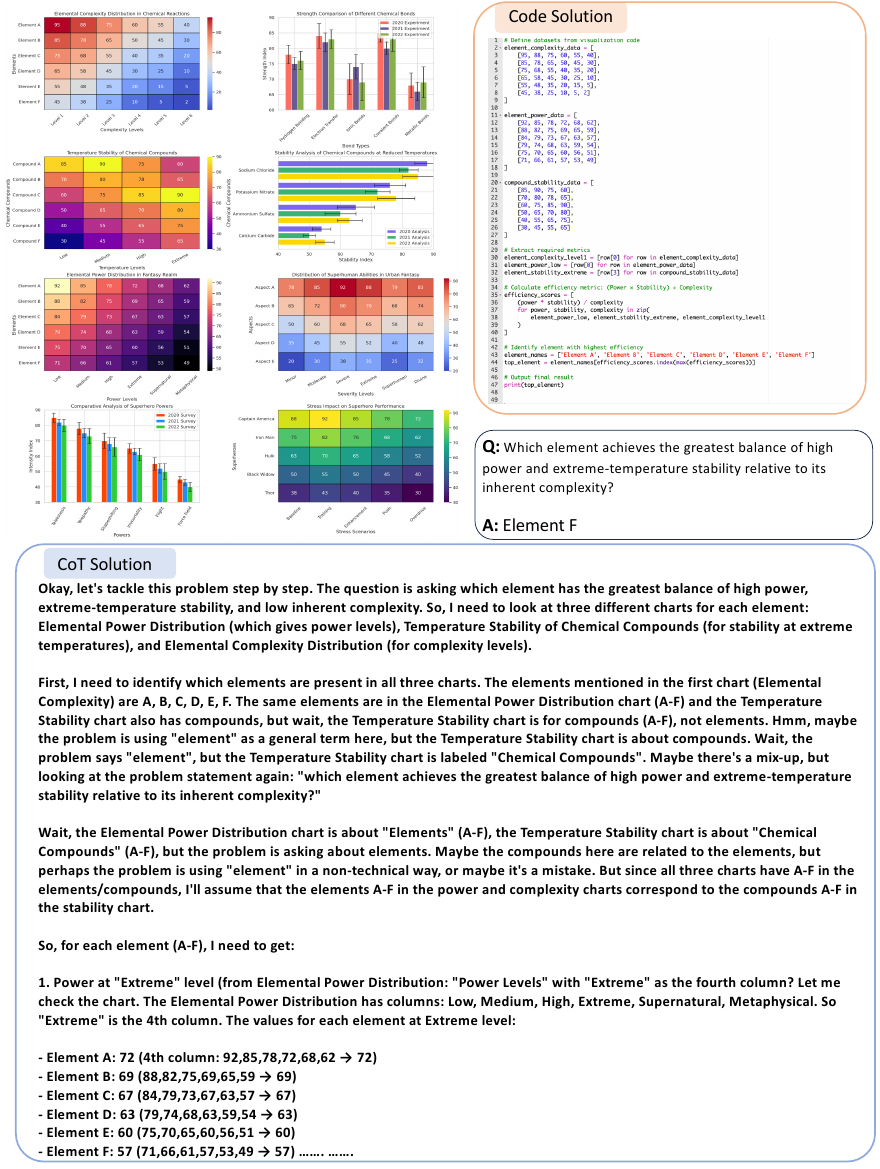}
    \caption{ChartVerse-SFT QA Example-3}
    \label{fig:qa_3}
\end{figure*}

Furthermore, we showcase the intricate QA pairs generated within our dataset through specific case studies shown in Figures~\ref{fig:qa_1}, \ref{fig:qa_2}, and \ref{fig:qa_3}. Leveraging the high information density of the synthesized charts, our pipeline is capable of formulating highly challenging queries. As evident in these examples, the questions transcend simple data retrieval by necessitating rigorous multi-step reasoning and cross-subplot integration.

For instance, the example in Figure~\ref{fig:qa_1} presents a multi-condition verification task where the model must sequentially validate three separate metrics including average scores and total efficiency across distinct experiment sub-panels. Similarly, Figure~\ref{fig:qa_2} demonstrates a demand for complex derived calculations. Here the model must visually extract statistical markers such as quartiles and medians to determine the region with the greatest proportional variation. This design ensures that the model must develop a holistic understanding of the visual context rather than relying on local pattern matching. Finally, Figure~\ref{fig:qa_3} requires the model to aggregate information from multiple spatially distinct plots. To identify the element with the optimal trade-off, the model must synthesize data regarding power levels, temperature stability, and complexity scores from three separate charts.

\begin{table*}[t!]
    \centering
    \caption{Prompt for Claude-4-Sonnet to generate code}
    \resizebox{1.0\textwidth}{!}{%
    \begin{tcolorbox}[colback=gray!5,colframe=black!75, title=Prompt for Claude-4-Sonnet to generate code]
    \small
    \#\# Role \\
    You are a matplotlib expert. Given a chart image, generate Python code that accurately reproduces this visualization.

    \#\# Inputs \\
    \textbf{Chart Image}: The visualization to reproduce

    \#\# Requirements
    \begin{enumerate}
        \item \textbf{Code Structure}:
        \begin{itemize}
            \item Import necessary libraries
            \item Include clear comments explaining the code only (no words like "reproduce", "match", "original", etc.)
            \item Ensure code is fully executable
            \item \textbf{DO NOT use read\_csv or any file reading operations}
            \item \textbf{Define all data directly in the code as variables/lists/dictionaries}
            \item \textbf{DO NOT use plt.show(), save the figure as 'image.png' instead}
        \end{itemize}

        \item \textbf{Visual Elements to Match}:
        \begin{itemize}
            \item Chart type and all data points
            \item Colors, styling, and proportions
            \item Title, axis labels, legend, and grid
        \end{itemize}

        \item \textbf{Text Positioning}:
        \begin{itemize}
            \item \textbf{Ensure all text labels are clearly visible and DO NOT overlap}
            \item \textbf{Include loc='best' parameter when setting legend to avoid text conflicts}
            \item Position each label carefully in its appropriate location
            \item Adjust text placement as needed to maintain readability
            \item Use \texttt{plt.tight\_layout()} to optimize overall spacing
        \end{itemize}

        \item \textbf{Output Format}:
        \begin{itemize}
            \item First briefly analyze the chart type and data structure.
            \item Then provide complete matplotlib code: \\
            \texttt{\`{}\`{}\`{}python} \\
            \texttt{[Code]} \\
            \texttt{\`{}\`{}\`{}}
        \end{itemize}
    \end{enumerate}

    \#\# Key Points
    \begin{itemize}
        \item \textbf{IMPORTANT: ensure data is reasonable(avoid nan)}
        \item Text labels must not overlap - adjust positions to keep them readable
        \item Match visual style as closely as possible
    \end{itemize}
    \end{tcolorbox}
    }
    \label{tab:claude_prompt}
\end{table*}
\begin{table*}[t!]
    \centering
    \caption{System Prompt for Complexity-Aware Chart Coder}
    \resizebox{1.0\textwidth}{!}{%
    \begin{tcolorbox}[colback=gray!5,colframe=black!75, title=System Prompt for Complexity-Aware Chart Coder]
    \small
    \#\# Role \\
    You are a Python visualization expert.

    \#\# Goal \\
    Generate a random Python visualization code focusing on charts, tables, or diagrams.

    \#\# Requirements
    \begin{enumerate}
        \item \textbf{Scope \& Tools}:
        \begin{itemize}
            \item Choose any visualization type (chart, table, flowchart, diagram, etc.).
            \item Create comprehensive sample data.
            \item Use standard Python visualization libraries (e.g., matplotlib, graphviz).
        \end{itemize}

        \item \textbf{Visual Design}:
        \begin{itemize}
            \item Ensure the visualization is visually appealing with proper labels, titles, and color schemes.
            \item Include sufficient visual elements to demonstrate complexity.
        \end{itemize}

        \item \textbf{Layout \& Clarity (Critical)}:
        \begin{itemize}
            \item \textbf{Carefully design the layout to avoid any overlapping text or elements.}
            \item Adjust figure size, margins, and spacing parameters for optimal clarity.
        \end{itemize}

        \item \textbf{Output Format}:
        \begin{itemize}
            \item Only output the Python visualization code wrapped in: \\
            \texttt{\`{}\`{}\`{}python} \\
            \texttt{[Code]} \\
            \texttt{\`{}\`{}\`{}}
        \end{itemize}
    \end{enumerate}
    \end{tcolorbox}
    }
    \label{tab:system_coder_prompt}
\end{table*}
\begin{table*}[t!]
    \centering
    \caption{Prompt for Inverse Logic Construction - Step 1: Logic Architecture \& Code Generation}
    \resizebox{1.0\textwidth}{!}{%
    \begin{tcolorbox}[colback=gray!5,colframe=black!75, title=Step 1: Logic Architecture \& Code Generation]
    \small
    \#\# Role \\
    You are an expert Data Scientist specializing in \textbf{Algorithmic Synthesis}. Your task is to look at raw data arrays from visualization code and engineer sophisticated, multi-step quantitative operations without needing instructions.

    \#\# Objective \\
    You will be provided with a \textbf{Visualization Code Snippet}. Your workflow is:
    \begin{enumerate}
        \item \textbf{Decode}: Extract the raw data signals from the visualization code.
        \item \textbf{Engineer}: Design a \textbf{multi-stage analysis} logic.
        \item \textbf{Implement}: Write a standalone Python script that executes this logic to produce a single significant \textbf{result (a specific number or a text label)}.
    \end{enumerate}

    \#\# Design Constraints (Difficulty Control)
    \begin{itemize}
        \item \textbf{Target Level}: The logic must be \textbf{accessible to a High School student}.
        \item \textbf{Linearity}: Focus on clear, linear logic, but avoid trivial operations.
    \end{itemize}

    \#\# Logic Architecture (The "Triangle of Complexity")
    You must invent a calculation path that adheres to these rules:
    \begin{enumerate}
        \item \textbf{Cross-Variable Correlation}: You must use interactions between at least three different data lists.
        \item \textbf{Multi-Subplot Integration (CRITICAL)}: If the input code contains multiple subplots, your calculation logic \textbf{MUST synthesize data across at least 2 subplots}.
        \item \textbf{Structural Dependency}: Do not simply process the dataset uniformly.
        \item \textbf{Derived Aggregation}: The final output must be a single synthesized value (numerical scalar OR categorical label).
    \end{enumerate}

    \#\# Output Requirements
    \begin{enumerate}
        \item \textbf{Data Context}: A brief professional summary of what the data represents.
        \item \textbf{Logic Blueprint}: A step-by-step text description of the algorithm.
        \item \textbf{The Python Script (The Solution)}:
        \begin{itemize}
            \item \textbf{Self-Contained}: Define all data lists explicitly inside the script.
            \item \textbf{Narrative Variable Naming}: Variable names \textbf{MUST} serve as documentation.
            \item \textbf{Atomic Output}: The script must end with exactly \textbf{ONE} \texttt{print()} statement showing \textbf{ONLY} the final result.
            \item Wrap the code in: \texttt{\textless answer\textgreater [Your Code] \textless /answer\textgreater}
        \end{itemize}
    \end{enumerate}

    \#\# Input Data \\
    \textbf{Input Code for Analysis:} \\
    \texttt{\{chart code\}}
    \end{tcolorbox}
    }
    \label{tab:script_prompt}
\end{table*}

\begin{table*}[t!]
    \centering
    \caption{Prompt for Inverse Logic Construction - Step 2: Semantic Reverse-Engineering}
    \resizebox{1.0\textwidth}{!}{%
    \begin{tcolorbox}[colback=gray!5,colframe=black!75, title=Step 2: Semantic Reverse-Engineering (Question Synthesis)]
    \small
    \#\# Role \\
    You are an expert in \textbf{Reverse-Engineering} complex code logic into natural, high-level business or scientific inquiries.

    \#\# Objective \\
    Your task is to formulate the \textbf{Analytical Question} that would prompt a human to write exactly the provided Solution Script based on the Original Visualization Code.

    \#\# Instructions

    \textbf{Phase 1: Logic Analysis} \\
    Analyze the "Solution Script" to understand \textit{what} is being calculated. Look at the variable names to determine the \textbf{INTENT}.

    \textbf{Phase 2: Question Formulation} \\
    Write the question that requires this exact logic to solve.

    \textbf{Constraint A: Explicit Target (Crucial)}
    \begin{itemize}
        \item The question must ask for a specific number OR a specific entity.
        \item \textbf{Avoid Vague Phrasing}: Ask for the specific metric calculated.
    \end{itemize}

    \textbf{Constraint B: Strict Logic Mapping}
    \begin{itemize}
        \item \textbf{The Code IS The Answer}: Ensure the provided Python script is the \textbf{exact, step-by-step solution} to your question.
    \end{itemize}

    \textbf{Constraint C: Semantic Abstraction (Most Important)}
    \begin{itemize}
        \item \textbf{PROHIBITION}: Do NOT describe mechanical steps (e.g., "Add list A and B").
        \item \textbf{REQUIREMENT}: Describe the \textbf{INTENT} and \textbf{DOMAIN MEANING} (e.g., "Calculate the profit margin").
    \end{itemize}

    \#\# Output Format
    Structure your response strictly as follows:
    \begin{enumerate}
        \item \textbf{Logic interpretation}: Briefly explain the semantic meaning of the Python script.
        \item \textbf{The Analytical Question}: The abstract, scenario-based question wrapped in \texttt{\textless question\textgreater ... \textless /question\textgreater}.
    \end{enumerate}

    \#\# Input Data \\
    \textbf{1. Original Visualization Code:} \\
    \texttt{\{chart code\}}

    \medskip
    \textbf{2. The Solution Script (The Answer Logic):} \\
    \texttt{\{generated\_python\_code\}}
    \end{tcolorbox}
    }
    \label{tab:qa_prompt}
\end{table*}

\begin{table*}[htbp]
    \centering
    \caption{Prompt for Inverse Logic Construction - Step 3: LLM Inference}
     \resizebox{1.0\textwidth}{!}{%
    \begin{tcolorbox}[colback=gray!5,colframe=black!75, title=Step 3: LLM Inference Prompt]
    \small
    \#\# Role \\
    You are an expert in \textbf{Chart Code Comprehension} and \textbf{Data Reasoning}. Your task is to answer a specific question by interpreting the raw data structures defined in a Python code snippet.

    \#\# Core Principles
    \begin{enumerate}
        \item \textbf{Code as Data Source}: Treat the Python code as a structured document containing the ground truth data. You do not need to "run" the code, but you must "read" and understand the variables (lists, dictionaries, values).
        \item \textbf{Semantic Mapping}: Map the terms in the question to the corresponding variables and indices in the code.
        \item \textbf{Logical Derivation}: Based on the extracted data, perform the necessary logical reasoning to answer the question.
    \end{enumerate}

    \#\# Solution Framework

    \textbf{Phase 1: Code Structure Analysis}
    \begin{itemize}
        \item Scan the code to identify key variables and data lists.
        \item Understand the relationship between different lists (e.g., \texttt{x\_axis} usually corresponds to \texttt{y\_axis} by index).
    \end{itemize}

    \textbf{Phase 2: Information Extraction \& Reasoning}
    \begin{itemize}
        \item \textbf{Locate Data}: Pinpoint the specific data points (values, labels, axis limits) in the code required by the question.
        \item \textbf{Step-by-Step Derivation}: Execute a clear, logical reasoning process. First, explicitly quote the values found in the code. Then, perform the necessary logical comparisons or arithmetic calculations step-by-step to derive the answer.
        \item \textbf{Note}: Rely ONLY on the information explicitly present in the code.
    \end{itemize}

    \textbf{Phase 3: Answer Formulation}
    \begin{itemize}
        \item Formulate a clear, concise answer based on your findings.
        \item Ensure the answer directly addresses the inquiry.
    \end{itemize}

    \#\# Strict Formatting Requirements
    \begin{enumerate}
        \item The final result goes ONLY inside \texttt{\textless answer\textgreater...\textless /answer\textgreater} tags.
        \item The last line of your response must be EXACTLY: "Therefore, the final answer is \texttt{\textless answer\textgreater ANSWER\textless /answer\textgreater}."
    \end{enumerate}

    \#\# Input Data \\
    \textbf{1. Chart Code:} \\
    \texttt{\{chart code\}}

    \medskip
    \textbf{2. Question:} \\
    \texttt{\{generated\_question\}}
    \end{tcolorbox}
    }
    \label{tab:consistency_prompt}
\end{table*}
\begin{table*}[t!]
    \centering
    \caption{Prompt for Vision-Language Model (VLM) CoT Distillation}
    \resizebox{1.0\textwidth}{!}{%
    \begin{tcolorbox}[colback=gray!5,colframe=black!75, title=CoT Distillation Prompt]
    \small
    \#\# Role \\
    You are an expert in science and visual reasoning with advanced capabilities in multimodal analysis. Your response will be used as a high-quality example to train a new AI model. Solve the problem efficiently and clearly by integrating \textbf{ALL} information from multimodal inputs.

    \#\# Core Principles
    \begin{enumerate}
        \item \textbf{Equal Weight to All Inputs}: Information from images is AS IMPORTANT as text. Never ignore visual elements.
        \item \textbf{Systematic Analysis}: Follow a rigorous, reproducible approach.
        \item \textbf{Precision and Accuracy}: Double-check all calculations.
        \item \textbf{Adaptive Reasoning}: Choose the most appropriate method based on context.
    \end{enumerate}

    \#\# Solution Framework

    \textbf{Phase 1: Comprehensive Information Extraction}
    \begin{itemize}
        \item Carefully analyze ALL text content for requirements and constraints.
        \item Thoroughly examine ALL visual elements, extracting every piece of relevant information.
        \item Explicitly connect visual and textual information.
    \end{itemize}

    \textbf{Phase 2: Strategic Problem Setup}
    \begin{itemize}
        \item Compile information and clearly state what needs to be found.
        \item Identify relevant principles and state assumptions.
    \end{itemize}

    \textbf{Phase 3: Rigorous Solution Execution}
    \begin{itemize}
        \item Present solution with complete logical flow and proper notation.
        \item Reference specific parts of visual inputs to support reasoning.
        \item Maintain unit consistency and precision.
    \end{itemize}

    \textbf{Phase 4: Solution Validation}
    \begin{itemize}
        \item Verify the answer makes scientific/logical sense.
        \item Ensure dimensional analysis is correct.
    \end{itemize}

    \#\# Strict Formatting Requirements
    \begin{enumerate}
        \item The final result goes ONLY inside \texttt{\textless answer\textgreater...\textless /answer\textgreater} tags.
        \item Include units inside the tags when required (e.g., \texttt{\textless answer\textgreater 5.2 m/s\textless /answer\textgreater}).
        \item The last line must be EXACTLY: "Therefore, the final answer is \texttt{\textless answer\textgreater ANSWER\textless /answer\textgreater}."
    \end{enumerate}

    \#\# Input Data \\
    \textbf{Problem:} \\
    \texttt{\{question\}} \\
    \textit{(Visual Inputs are provided via the vLLM interface)}
    \end{tcolorbox}
    }
    \label{tab:distill_prompt}
\end{table*}



\end{document}